\documentclass[lettersize,journal]{IEEEtran}
\usepackage{amsmath,amsfonts}
\usepackage{algorithm}
\usepackage{algpseudocode}
\usepackage{array}
\usepackage{textcomp}
\usepackage{stfloats}
\usepackage{url}
\usepackage{verbatim}
\usepackage{graphicx}
\usepackage{booktabs}
\usepackage{cite}
\usepackage{xcolor}
\usepackage{soul}
\usepackage{multirow} 
\usepackage{fancyhdr}
\usepackage{tabularx}  
\usepackage{makecell} 
\usepackage{hyperref}
\usepackage[caption=false,font=normalsize,labelfont=sf,textfont=sf,caption=false,font=footnotesize]{subfig}
\begin{document}

\title{Long-Horizon Traffic Forecasting via Incident-Aware Conformal Spatio-Temporal Transformers}

% \author{Mayur Patil, Qadeer Ahmed, Shawn Midlam-Mohler, Stephanie Marik, Allen Sheldon, Rajeev Chhajer, Nithin Santhanam
% % \thanks{The authors are with the Department of Mechanical and Aerospace Engineering and the Center for Automotive Research, The Ohio State University, Columbus, OH 43212 USA (e-mail: patil.151@osu.edu; ahmed.358@osu.edu; midlam-mohler.1@osu.edu)}% <-this % stops a space
% \thanks{The authors are with the Department of Mechanical and Aerospace Engineering, (Center for Automotive Research, The Ohio State University), Columbus, OH 43212 USA (e-mail: patil.151@osu.edu; ahmed.358@osu.edu; midlam-mohler.1@osu.edu)}}

\author{
Mayur Patil$^{1}$, Qadeer Ahmed$^{1}$, Shawn Midlam-Mohler$^{1}$, Stephanie Marik$^{2}$, Allen Sheldon$^{3}$, Rajeev Chhajer$^{3}$, Nithin Santhanam$^{3}$%
\thanks{$^{1}$Mayur Patil, Qadeer Ahmed, and Shawn Midlam-Mohler are with the Department of Mechanical and Aerospace Engineering and the Center for Automotive Research, The Ohio State University (e-mail: \texttt{patil.151@osu.edu}; \texttt{ahmed.358@osu.edu}; \texttt{midlam-mohler.1@osu.edu}).}%
\thanks{$^{2}$Stephanie Marik is with the Ohio Department of Transportation}%
\thanks{$^{3}$Allen Sheldon, Rajeev Chhajer, and Nithin Santhanam are with Honda Research Institute USA, Inc.}%
}

\pagestyle{fancy}
\fancyhead{} 
% \fancyfoot{} 
% \fancyfoot[C]{\color{red}\footnotesize This manuscript has been accepted as a REGULAR PAPER in the Transactions on Intelligent Transportation Systems 2025}
\renewcommand{\headrulewidth}{0pt} 

\maketitle
\thispagestyle{fancy}


\maketitle

\begin{abstract} % Mention that this is done on real data...SUMO line...crash data is comign from ODOT
Reliable multi-horizon traffic forecasting is challenging because network conditions are stochastic, incident disruptions are intermittent, and effective spatial dependencies vary across time-of-day patterns. This study is conducted on the Ohio Department of Transportation (ODOT) traffic count data and corresponding ODOT crash records. This work utilizes a Spatio-Temporal Transformer (STT) model with Adaptive Conformal Prediction (ACP) to produce multi-horizon forecasts with calibrated uncertainty. We propose a piecewise Coefficient of Variation (CV) strategy that models hour-to-hour travel-time variability using a log-normal distribution, enabling the construction of a per-hour dynamic adjacency matrix. We further perturb edge weights using incident-related severity signals derived from the ODOT crash dataset that comprises incident clearance time, weather conditions, speed violations, work zones, and roadway functional class, to capture localized disruptions and peak/off-peak transitions. This dynamic graph construction replaces a fixed-CV assumption and better represents changing traffic conditions within the forecast window. For validation, we generate extended trips via multi-hour loop runs on the Columbus, Ohio, network in SUMO simulations and apply a Monte Carlo simulation to obtain travel-time distributions for a Vehicle Under Test (VUT). Experiments demonstrate improved long-horizon accuracy and well-calibrated prediction intervals compared to other baseline methods.
\end{abstract}


\section{Introduction}
\IEEEPARstart{U}{rban} traffic networks operate under continual incidental changes such as ever-changing weather, road constructions, breakdowns, crashes, special events, and heterogeneous driver behaviors. Due to these variables, forecasting models must adapt to changes that can occur within short time periods throughout the day. According to the 2024 INRIX Global Traffic Scorecard \cite{INRIX2024}, an average driver in the U.S. spends about 43 hours stuck in traffic, resulting in an estimated loss of \$771 in time for each driver. New York and Chicago ranked the highest, with drivers in those cities losing 102 hours each, highlighting the severity of the issue. At the same time, roadway safety remains a major concern; the most recent national crash totals are reported for 2023 (6.14 million police-reported crashes, 40,901 fatalities, and about 2.44 million injuries) \cite{NHTSA2023Crashes}. Full-year crash totals for 2024 are not yet finalized; however, NHTSA's early estimates indicate that about 39,345 people died in motor vehicle traffic crashes in 2024 \cite{NHTSA2024}. These numbers motivate predictive models that are aware of incident-driven disruptions in traffic operations, beyond recurrent congestion patterns. Weather further complicates operations by amplifying delays and risks during peak periods. Federal guidance indicates that adverse weather is the second-largest source of non-recurrent congestion, where even light rain can inflate travel-time delays by 12-20\% \cite{FHWAWeather}. The practical need is clear: agencies and mobility providers require reliable, long-horizon forecasts with calibrated confidence to plan operations under variability and disruptions.

From a modeling standpoint, long-horizon/multi-hour traffic forecasting has always been a challenge as traffic in itself evolves throughout the day (off-peak to peak hour and back). Furthermore, exogenous factors such as roadway incidents, weather, and construction work disrupt traffic flow patterns. Conventional sequence models, such as recurrent neural networks (RNNs), are capable of handling temporal features, but they often run into issues with error accumulation and have less freedom to perform parallel processing over longer time periods. While the transformer-based designs use a self-attention mechanism to find long-range temporal relationships, and can be adjusted to include spatial relationships. Recent developments in transportation applications show that the attention-based approach can improve multi-hour accuracy and is capable of error stabilization by explicitly learning which time steps and locations matter most as the horizon extends \cite{STTN,Cai2020,Yan2022}.

The purpose of leveraging a transformer model for long-horizon forecasting is to look far back in history without encountering the vanishing gradient problem. This provides an adaptable mechanism to share the information spatially and scale it efficiently using parallel attention. Simultaneously, long-horizon accuracy significantly depends on how the spatial dependencies are represented, as the information of the locations can vary with time-of-day variability and disruptions caused by exogenous incidents.

\begin{figure}[!t]
    \centering
    \includegraphics[width=\linewidth]{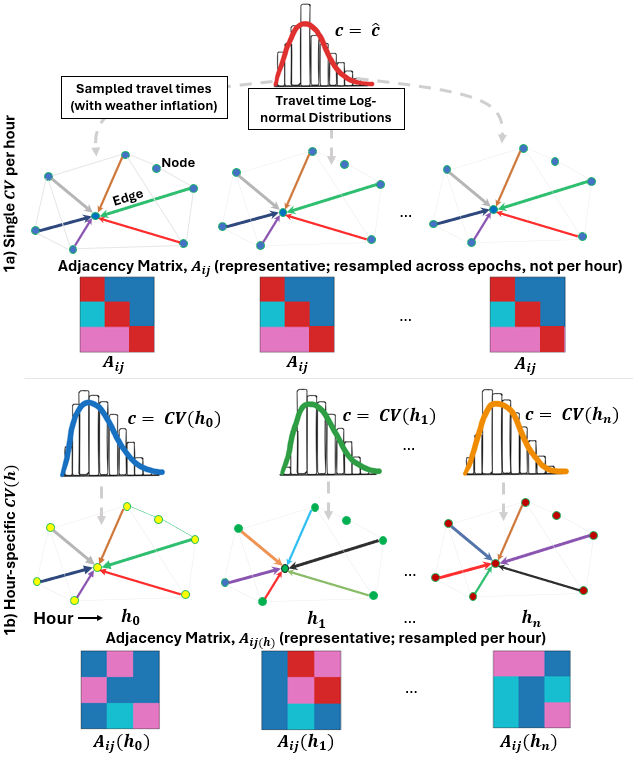}
    \caption{Static vs.\ piece-wise graph priors: 1a) one fixed-CV $\hat c$ drives a log-normal travel-time prior; the resulting adjacency $A_{ij}$ is reused across hours, 1b) an hourly profile $CV(h)$ yields hour-conditioned priors and distinct adjacencies $A_{ij}(h)$. Edge colors/thicknesses are schematic.}
    \label{fig:intro}
\end{figure}

In our earlier work \cite{Patil2025}, we leveraged a Graph Attention Network with Long-Short Term Memory (GAT-LSTM) with Adaptive Conformal Prediction (ACP), constructing the spatial adjacency matrix $A_{ij}$ from a log-normal distribution representing travel-time information. The log-normal distribution are widely used to represent travel-time stochasticity and has been reported to fit observed travel times across operating regimes in both roadway and transit settings \cite{Mazloumi2010,Shen2019}. We parameterized the distribution using a fixed coefficient of variation (CV) by taking random values of travel-time to make $A_{ij}$ dynamic in nature. Basically, it governed the dispersion of sampled travel times that we mapped onto the edge weights of GAT, and the same graph effectively governed all hours as shown in Figure~\ref{fig:intro}\,a. Although practically on average, the fixed-CV assumption implicitly treats the network's variability as time-invariant.

In this work, we replace the ``fixed-CV" assumption with a piecewise design by estimating an hour-of-day CV profile $CV(h)$ from historical traffic flow data (higher CV represents peak times, while lower CV represents off-peak/shoulder times). The intent is to sample log-normal edge travel times for each hour $h$ using $CV(h)$, then introduce an incident-aware perturbation using a crash-derived severity signal calculated using incident clearance time, weather factors, speed-violation, work zones, and roadway functional class. To set up the correlation scale, we estimate edge-wise correlations between the sampled travel-time and the crash-derived attributes. We then modulate the adjusted travel times into hour-conditioned adjacency weights $A_{ij}(h)$ through a decaying kernel and normalization. This $A_{ij}(h)$ is then a set of 24-hour-conditioned graphs $A_{ij}(h)$ being fed to the model, where each training window carries its hour tag and applies the corresponding graph as visualized in Figure ~\ref{fig:intro}\,b. To quantify uncertainty in our predictions, we retain the ACP methodology, leveraging the conformal calibration core to provide reliable confidence bounds.

To accomplish this task, we adopt a Spatio-Temporal Transformer (STT) architecture for efficient, parallel sequence modeling and integrate hour-conditioned graph derived from the proposed hour-conditioned adjacency matrices. The contributions of this paper are as follows:
\begin{itemize}
    \item We consider an hourly CV profile from historical traffic flow data, where each hour's CV parameterizes a log-normal distribution sample of travel times. This yields a set of hour-conditioned adjacency matrices $A_{ij}(h)$ for multi-hour forecasting.
    \item We introduce crash-aware perturbations of edge weights using incident-related severity signals derived from a crash dataset (incident clearance, weather factors, work zone areas, speed violations, and roadway functional class), enabling localized disruption sensitivity in the dynamic graph.
    \item We retain ACP for finite-sample coverage guarantees on forecasts, yielding reliable confidence around forecasted values.
    \item We run multi-hour SUMO simulations via Monte Carlo sampling for a Vehicle-Under-Test (VUT) trajectory and validate it against INRIX travel-time data.
\end{itemize}

The rest of the paper is organized as follows: Section II summarizes prior work on traffic flow forecasting. Section III presents the proposed framework and methodology. Section IV reports the experimental setup and comparative results against baseline models. Section V concludes the paper and outlines future research directions.

% \IEEEPARstart{T}{raffic} 

\section{Related Work} % Cite trasportation research part C papers!
\subsection{Traffic Forecasting Models and Graph Learning}
In the past, traffic forecasting was conducted using conventional univariate and multivariate time-series models that considered each sensor stream independently. Because of its simplicity and interpretability, the original Box-Jenkins family of AutoRegressive Integrated Moving Averages (ARIMA) continues to be a typical baseline for short-term forecasting \cite{BoxJenkins}. Similarly, exponential-smoothing methods such as Simple Exponential Smoothing (SES) \cite{Brown1963}, Holt's linear trend method \cite{Holt1957}, Holt-Winters seasonality \cite{Winters1960}, and Error Trend-Seasonality (ETS) state-space models \cite{HyndmanETS} are other short-term traffic speed/flow prediction \cite{Lippi2013}. However, these approaches are only limited to short-horizon prediction as their linear dynamics struggle with nonlinear congestion patterns, they do not have encoding for spatial interactions, and their mechanism is restrictive to handle disruptions such as road incidents, demand surges, weather, etc.

However, the field made a significant impact when deep sequence models were introduced. They were shown to be capable of learning nonlinear dynamics inherently from the data. The early stacked feedforward and autoencoder-based models showed good performance for short-term traffic forecasting \cite{Lv2015}. Recurrent neural networks (RNNs) like LSTM/GRU variants improved multi-step prediction by maintaining temporal state and reducing the vanishing gradients problem compared to vanilla RNNs \cite{Ma2015,Fu2016}. Long-horizon scenarios, however, may result in the accumulation of error due to repeated rollouts, and unless explicitly introduced by design, spatial relationships remain implicit.

The bottleneck for long-horizon forecasting is that the model has to decide ``what to remember" from the history and ``how to roll forward" without ``losing attention". First and foremost, many temporal modeling approaches were proposed specifically to reduce this distraction. For instance, Seq2Seq encoder-decoder models addressed this problem by separating historical encoding from future decoding, so that the forecasting depends on a concise representation of the historical data and have an option to attend back to it when needed. Similarly. convolution-based approaches like LSTNet combines temporal convolutions and can skip certain connections to capture important patterns at \cite{Lai2018LSTNet}, whereas Temporal Convolution Network (TCN) use dilated causal convolutions to model long-horizon aspects without distinct recurrent states \cite{Bai2018TCN}. Furthermore, attention-augmented recurrent models like dual-stage attention-based recurrent neural network (DA-RNN) modify the weights based on the current context, hence handling the off-peak and peak condition changes \cite{Qin2017DARNN}. Secondly, the spatio-temporal forecasting models improved when they coupled the temporal model with an explicit graph-based framework to share information spatially. The key idea is that the model should be able to pass messages between correlated road segments rather than expecting the temporal module to infer spatial relationships. The early baselines implemented this by integrating temporal module with graph convolutions on a graph prior. For instance, STGCN fuses temporal convolutions with graph convolutions to capture general space-time relationships \cite{STGCN}. Then, DCRNN approach proposed a diffusion process on a directed graph and proposed a gated recurrence to modify node states over time \cite{DCRNN}, while T-GCN approach similarly integrates convolution-based spatial integration with gated temporal dynamics \cite{TGCN}. Furthermore, an attention-based extension was put forth by ASTGCN model which relaxed rigid neighborhood influence by learning spatial-temporal important events, thus capturing periodic information (hourly/daily/weekly) and heterogeneous node impacts \cite{ASTGCN}. Beyond these common baselines, many variants refines the temporal module, graph structure, or both. For example, traffic graph convolutional recurrent designs bolster the prediction framework by explicitly communicating information along the graph edges rather than over-utilizing the temporal module to re-learn spatial information from scratch \cite{TGCLSTM}.

Another advancement was made, and spatio-temporal GNNs were matured in terms of introducing dynamics in creating the adjacency matrix. Methods were proposed to make a shift from a static distance-based adjacency matrix to a more adaptive adjacency that is learned from data. Graph WaveNet introduced a data-driven spatial module and created the adjacency matrix alongside dilated causal convolutions, greatly reducing error accumulation across multi-step horizons in the case of incomplete/misspecified sensor information \cite{Wavenet}. Other models, such as GMAN used an encoder-decoder attention architecture to learn when and where to accumulate the information, while replacing the custom-built diffusion layer with learned spatio-temporal aggregation \cite{GMAN}. A parallel line of work studied explicitly dynamic graphs, e.g., dynamic graph convolution formulations for multi-step traffic forecasting \cite{DGCN}, multi-weight graph architecture that put together multiple information types \cite{MWTGC}, and graph-learning STGCN variants that utilized the adjacency matrix as part of training data \cite{GLSTGCN}. Other work emphasized on fusing various features/attributes, where multiple other factors of the network state were integrated instead relying on a single traffic attribute to improve robustness under dynamic operating conditions \cite{MFFB}. Interestingly, meta-learning has also been explored to improve transferability across the network, e.g., spatio-temporal meta-learning frameworks being capable of generalizing under distribution shifts \cite{STMetaNet}, and multivariate time-series graph forecasting models such as MTGNN proposed on learning graph-structure and message passing specific to high-dimensional correlated series \cite{MTGNN}.

Despite significant progress, most of the above studies frequently considered short-horizon forecasting (typically 15-60 minutes) and/or under very smooth connectivity assumptions. On the other hand, multi-hour forecasting (e.g., predicting 2, 4, 8, or 12 hours ahead) is harder because the spatial influence pattern is not time-invariant across the day as there is peak build-up, peak dissipation, and localized disruptions that affect which links dominate the changes. This is precisely the gap we target and prioritize multi-hour forecasting by providing an hour-conditioned, variability-aware spatial information instead of a single static graph, so that long-horizon performance is not driven purely by extrapolation from a stationary connectivity assumption.

\subsection{Long-horizon forecasting}
Only a small but rising number of papers worked on the long-horizon forecasting problem. An interesting Graph Pyramid Autoformer family (X-GPA/GPA) was proposed to explicitly evaluate multi-hour settings (e.g., 2/4/8/12 hours). To focus on long-range shift stabilization, an autocorrelation-based attention approach was utilized with a pyramidal multi-scale temporal representation ~\cite{Zhong2022XGPA,Zhong2023GPA}. Similarly, long-term graph learning studies also worked in the same avenue and support the claim that extending forecasting horizon beyond several hours requires stronger inductive bias to prevent model drift, including formulations that explicitly evaluate long-range prediction on graphs~\cite{Bogaerts2020GraphCNNLSTM,Yu2022LongTermGraphs}.

The latest works push the design capability and extend the prediction horizons. One such model, TrafFormer, specifically defines the task of predicting traffic up to 24 hours using a Transformer framework tailored for handling long-term sequences ~\cite{TrafFormer2023}. Another model, HUTFormer, similarly does long-term forecasting beyond 1-hour and discusses the outlook on up to 1 day by leveraging hierarchical multi-scale representations to handle long input/output horizons comprehensively ~\cite{HUTFormer2023}. Additionally, system-oriented work such as Foresight Plus focuses on establishing spatio-temporal traffic forecasting in a serverless scenario. Also, it discusses practical approaches for running the latest long-horizon models in a production pipeline~\cite{ForesightPlus2024}.

Another approach uses common traffic baselines to assess robustness in multi-hour regimes explicitly. Using roughly 2-hour and 4-hour prediction settings on 5-minute datasets (e.g., PEMS-BAY), SpikeSTAG reports long-horizon results, emphasizing that multi-hour horizons reveal vulnerabilities in short-horizon-tuned architectures and that performance gains are frequently linked to maintaining cross-node propagation structure as the horizon expands ~\cite{Hu2025SpikeSTAG}. Lastly, a GAT-Informer model was proposed that leverages a Transformer-like long-term forecasting module with graph attention and an Informer-like temporal modeling to perform tests at long horizons (e.g., up to 2 hours on a 15-minute dataset) ~\cite{Song2024GATInformer}.

\subsection{Transformer Architectures for Traffic Forecasting}
An astounding innovation of self-attention was recently developed, and it took over the artificial intelligence field with incredible results. This was the wave of Transformers, which showed that instead of relying on recurrent cells or fixed-receptive-field convolutions, the architecture was built on the standard Transformer design \cite{Vaswani2017Attention} and uses multi-head attention to connect distant time steps and locations in a single layer. Hence, making them especially attractive for multi-hour horizons and highly non-stationary urban regimes. 

A Traffic Transformer approach \cite{Cai2020TrafficTransformer} was one of the first traffic-specific Transformer designs, which extensively used the continuity and periodicity of traffic time series. The architecture simultaneously models spatial correlations between the traffic stations and incorporates temporal aspects into an encoder module. Large-scale urban dataset experiments clearly outperform RNN/CNN baselines, especially when long-range periodic patterns are critical. On the same lines, Spatio–Temporal Transformer Networks (STTN) model was proposed, which decoupled spatial and temporal modules into a separate Transformer module that handled various traffic sensor graphs \cite{Xu2020STTN}. Here, the spatial module captures dynamic multi-head dependencies between the nodes, and the temporal module handles the long-range time-series history. This encoder–decoder design showed significant state-of-the-art performance on the widely used PeMS dataset, particularly at long horizons where small temporal modules and static graphs showed inadequate performance.

A dynamic, hierarchical spatial–temporal Transformer formulation was also proposed \cite{Yan2022DynamicHierarchical}, by stacking Transformer blocks with hierarchical attention patterns to capture short-term interactions (local) and long-term congestion (global) evolution simultaneously. The outcome was an architecture that outperforms STGCN/DCRNN baselines on multi-step traffic speed prediction and also provided interpretable attention maps showing roads and time periods influencing the performance. A bidirectional spatial–temporal adaptive Transformer (Bi-STAT) was later proposed by applying self-attention in both forward and backward temporal directions. Interestingly, they made the spatial attention pattern completely adaptive instead of a fixed adjacency matrix \cite{Chen2022BiSTAT}. This particularly raised interest in terms of how and why integration of incidents, lane closures, and other regime shifts that which is critical to capture real-world representation.

A number of studies focused on improving the temporal module of the Transformer architecture for traffic flow prediction. Specifically for traffic flow states, a Transformer-based deep neural network was proposed that enhanced the fundamental self-attention process, enhancing embedding and decoder structure \cite{Wen2023TransDeepTFP}. To better capture complex spatio–temporal correlations while maintaining efficiency, a Fast Pure Transformer Network (FPTN) was proposed that used sequential traffic data and introduced multiple embeddings for sensor position, and temporal/positional aspects \cite{Zhang2023FPTN}. An improved version of Transformer for traffic flow prediction was later proposed by modifying input embedding and temporal attention by explicitly modeling both long-range and short-range dependencies \cite{Liu2025ImprovedTransTFP}. Then, a Transformer-based short-term traffic forecasting model was introduced that systematically compared attention-based designs with conventional deep networks, while confirming that self-attention can produce more stable performance across different congestion levels and short horizons \cite{Chang2025ShortTermTrans}.

Transformers and graph-based spatial modeling are closely integrated in one of the other studies \cite{Ma2024PLMSTT}, introducing a spatial–temporal Transformer network that improves the traffic flow prediction using a pre-trained language model; introducing contextual information using Transformer blocks along with spatio–temporal inputs. A spatial–temporal Transformer-based traffic flow prediction model, ST-TransNet, was another distinct model for bridge networks, which jointly encodes correlations among bridges and temporal dynamics using stacked self-attention layers \cite{Tian2025STTransNet}. Furthermore, a graph-enhanced spatio–temporal Transformer was introduced that integrated graph convolutions with a Transformer layer to learn spatial representations and attention-based temporal modeling \cite{Kong2025GESTT}. Another approach took this further by separating graph-based spatial encoding and Transformer-based temporal encoding, and then combining them again \cite{Sun2025DGSTTN}. During the same time, a hybrid model was proposed that incorporated a spatio–temporal Transformer with graph convolutional networks (GCNs), exhibiting that joint attentions are much better in capturing regional interactions along with temporal dependencies \cite{Zhang2025STTGCN}. As multiple works are being done following the Transformer-based approach, the Transformer-Enhanced Adaptive Graph Convolutional Network (TEA-GCN) was introduced, where Transformer modules enhance the adaptive graph learning, allowing the spatial framework to be refined in a data-driven way during training \cite{He2024TEAGCN}. At a more comprehensive scale, a learnable long-range graph Transformer (LLGformer) was published that improved traditional Transformer-style models with techniques intended for more efficient learning of long-horizon dependencies in traffic flow data on large graphs \cite{Zhang2025LLGformer}. 

In the above studies, we found three common points: i) Multi-head self-attention temporally offers a flexible mechanism to extract longer/multi-scale histories, reducing the vanishing gradient problem and the limiting long-range training difficulties of RNNs and TCNs, ii) To allow cross-link effects to change with time and traffic conditions rather than being fixed by a static graph, many topologies changed from fixed, hand-crafted adjacency matrices to data-driven or adaptive spatial attention, and iii) In multi-step rollouts, encoder-decoder designs and partially parallel decoding techniques reduce exposure bias and error accumulation. However, the majority of the current works still consider the spatial structures relatively slowly varying, and they rarely encode variability caused by exogenous factors directly into the spatial module. In this work, we leverage the Transformer-based framework but fuse the encoder–decoder architecture with a variability-aware, piecewise, hour-conditioned graph derived from coefficient of variation choices. By injecting this connectivity into the Transformer architecture, the model has stochasticity available from the adaptive adjacency matrix purely from a data-driven source.

\subsection{Incident context features}
Non-recurrent congestion caused by crashes is a notable source of forecasting error whether it's short-term or long-term. Its consequences are majorly dependent on factors like the crash/incident clearance time, weather/road condition at the time of crash, the type of roadway (functional class), construction or work-zone disruption, speed infractions, so on and so forth \cite{FHWAIncidentClearance}. As crash incident response and recovery time can last from minutes to hours, the effects are especially appropriate for multi-hour horizons where the traffic states can transition from off-peak to peak hours to disrupted conditions in a single forecast window.

As a result, a substantial amount of research directly predicts the event dynamics, mostly through the prediction of incident length. The statistical learning and machine learning techniques such as hazard models, tree-based models, mixture models, and deep learning architectures can predict the length of the events and quantify the impact of other related factors as shown by some latest studies and review works \cite{Li2018IncidentReview,Korkmaz2024IncidentDurationReview,Corbally2024IncidentDuration}. These studies repeatedly emphasize that characteristics that control how disruptions develop, such as response situations and clearance-related variables, that influence incident impacts in addition to the event itself. This body of research encourages handling incident-related data as a prime signal rather than a rare incident that a general forecasting model should automatically consider.

Simultaneously, rich models for crash frequency, crash risk, and collision severity have been constructed by road-safety analytics. These models explicitly use exogenous descriptors, such as weather, roadway class/type, construction zones and variables related to speed. The occurrence and impact of crashes are structured in both location and time, as shown by spatio-temporal crash modeling and risk mapping \cite{Chaudhuri2022CrashST,Jin2024CrashPatterns}. Weather and highway context (including road hierarchy/functional class) are among the most relevant factors for explaining the crash outcomes, according to severity-prediction studies \cite{Savolainen2011SeverityReview,Penmetsa2018RoadFeature,Jamal2021EnsembleSeverity,Yan2021AdverseWeatherSeverity}. The factors of interest are majorly speed-related as NHTSA reports that over-speeding contributed almost 29\% of traffic fatalities in 2023, emphasizing why speed-violation indicators are important factors when characterizing crash severity \cite{NHTSASpeeding2023}. Transportation officials have long stressed that bad weather exacerbates both safety and mobility problems, as we pointed out that travel time is increased by 12-20\% \cite{FHWAWeather}, So, adverse weather is a leading contributor to non-recurrent delays.

In order to improve traffic speed prediction during disruptions, \cite{Xie2020DIGCNet} employs event detection and representation learning to convert raw incident information into latent features. These factors are frequently integrated with graph-based models to capture the propagation patterns. Although these models confirm that incident factors lowers error during disruption periods, incident information is typically incorporated as an additional inputs in the form of a time-series data, or learned embeddings rather than as a coupling them explicitly with the spatial dependency structure.

In this study, we utilize crash context features with dynamic spatial structure. We employ incident clearance time, weather, speed violations, work zone areas and functional class to construct an incident-severity signal that modifies edge weights in an hour-conditioned adjacency matrix. This approach aligns with the logic found in the above studies where disruption impact depends on contextual factors, and its network influence is reflected in the effective connectivity used for prediction. By embedding this disruption signals into the spatial module itself, the forecasting model begins each window with a connectivity that is already biased towards the propagation patterns, rather than learning all such effects purely from historical time series.

\subsection{Uncertainty Quantification}
Uncertainty quantification is critical for multi-hour traffic forecasting, as uncertainty is state-dependent and increases with the horizon window. Essentially, crash-related disruptions (and their associated context, such as weather, speed-related behavior, work zone, and roadway functional class) can introduce abrupt distribution shifts that are not well represented by point forecasts alone. Consequently, systems do require calibrated confidence bounds in addition to point estimates. Earlier work in traffic forecasting has quantified uncertainty using Bayesian or approximate-Bayesian methods, quantile-based models, and resampling or ensemble strategies. Bayesian learning formulations have been applied to traffic speed prediction with uncertainty estimates \cite{WuYu2021BayesianUQ}. Other practical approaches, such as Monte Carlo dropout \cite{GalGhahramani2016Dropout} and deep ensembles \cite{Lakshminarayanan2017DeepEnsembles}, are widely used because they can utilize existing neural forecasters but at the cost of increased computation. Hence, they struggle to provide coverage guarantees under a distribution shift. 

To handle the distribution shifts, Conformal prediction (CP) offers an exceptionally good performance by offering a complementary, distribution-free route by calibrating interval width from validation residuals, resulting in finite-sample coverage guarantees \cite{Stankeviciute2021ConformalTS}. The CP family has conformalized quantile regression (CQR) that improves efficiency by conformalizing learned quantiles \cite{Romano2019CQR}, then for non-stationary settings, adaptive conformal methods were developed to update calibration online to maintain coverage under drift \cite{Gibbs2021ACI,Zaffran2022ACPTS}. In this work, we retain an adaptive CP (ACP) style conformal calibration layer while strengthening the base forecaster with crash-informed, time-varying graph connectivity. This pairing is adopted for incident-driven traffic where the dynamic graph reduces the misspecification during disruptions, while ACP provides reliability-controlled uncertainty intervals for the residual uncertainty.

% -------------------------------------------------------------------

\section{Proposed Methodology}
\subsection{Graph Formulation and Data Availability}
Following our prior formulation~\cite{Patil2025}, we model the sensor network as a directed graph $\mathcal{G}=(\mathcal{V},\mathcal{E})$ (nodes are stations; edges are feasible directed links).  where we have two types of stations i)  Continuous Count Stations (CCS), and ii) Non-Continuous Count Stations (N-CCS), providing traffic data in a rich (5/15 min interval; each day of the year) and sparse manner (maybe for a few days/months only), respectively. We are reiterating the approach in this work- each node $i$ is assigned a data-availability score $a_i\in[0,1]$:
\begin{equation}
a_i =
\begin{cases}
1, & i\ \text{is CCS}\\[3pt]
\displaystyle\frac{C_i}{\max_j C_j}, & i\ \text{is N-CCS}
\end{cases}
\label{eq:avail_node}
\end{equation}
where $C_i$ is the count data from N-CCS node $i$, and $max_jCj$ is the maximum count observed among all N-CCS stations. Then, we modify it to an edge-wise reliability mask as:
\begin{equation}
\mathbf{A}_{\text{avail}}[i,j]=a_i a_j
\label{eq:avail_outer}
\end{equation}
We apply $\mathbf{A}_{\text{avail}}$ as a multiplicative mask on the learned hour-conditioned adjacency in later sections. \textit{(Please note that $\mathcal{E}$ is the edge notation different than $\varepsilon>0$ which is used throughout the paper as tiny constant to avoid division by zero.)}

% -------------------------------------------------------------------

\subsection{Hourly Adaptive Adjacency Matrix Formulation}
\label{subsec:cv_sampling}

Considering network variability as time-variant is one of the main objectives of this work. Practically, the spread of travel times increases significantly during peak hours and decreases during off-peak hours. To capture this effect, we construct an hour-of-day coefficient-of-variation profile:
\begin{equation}
\text{CV}(h) \in [0.1, 1.0,], \qquad h \in \{0,1,\dots,23\}
\end{equation}
% estimated from historical flow data. Intuitively, small $\text{CV}(h)$ represents stable conditions (e.g., late night), while larger $\text{CV}(h)$ corresponds to highly variable periods (e.g., morning/evening peaks). In practice, we first compute an empirical coefficient of variation for each hour-of-day and then map it to a bounded range
estimated from historical flow data. Understandably, the low traffic conditions, like late night, are represented by a small CV(h), and heavy traffic times, like morning/evening peaks, are represented by a larger CV(h). For instance, if we consider a typical case, very early-morning hours such as 03:00-05:00 AM typically fall near $\text{CV}(h) \approx 0.15$-$0.25$ (free-flow with small spread), he mid-morning and early-afternoon periods around 08:00-10:00 and 13:00-14:00 rise to $\text{CV}(h) \approx 0.35$-$0.50$, and the main peaks around 16:00-18:00 can reach $\text{CV}(h) \approx 0.70$-$0.90$. After the evening peak, the coefficient gradually drops again, with late-night hours (21:00-01:00) returning toward $\text{CV}(h) \approx 0.20$-$0.30$. 

Let the baseline mean travel time between stations $i$ and $j$ be represented by $T^{\text{mean}}_{ij}$. For a given hour $h$, we model the stochastic travel time $T^{(h)}_{ij}$ as a log-normal distribution:
\begin{equation}
T^{(h)}_{ij} \sim \text{LogNormal}\!\big(\mu^{(h)}_{ij,\ln},\,\sigma^{(h)}_{\ln}\big)
\label{eq:logn_model}
\end{equation}
with the constraint that the distribution has mean $T^{\text{mean}}_{ij}$ and coefficient of variation $\text{CV}(h)$. For hour-specific cases, we compute the parameters of the lognormal distribution as:
\begin{equation}
\sigma^{(h)}_{\ln} \;=\; \sqrt{\ln\!\big(\text{CV}(h)^2 + 1\big)}
\label{eq:sigma_ln}
\end{equation}

\begin{equation}
\mu^{(h)}_{ij,\ln} \;=\; \ln\!\big(T^{\text{mean}}_{ij}\big)
\;-\; \frac{1}{2}\big(\sigma^{(h)}_{\ln}\big)^2
\label{eq:mu_ln}
\end{equation}

We then draw samples from $T^{(h)}_{ij}$ and assemble them into an hour-specific travel-time matrix, and repeating this for all $h\in\{0,\dots,23\}$ produces a piecewise (hour-indexed) collection of travel-time adjacency matrices:
\begin{equation}
\big\{\mathbf{T}^{(h)}\big\}_{h=0}^{23}
\label{eq:T_bank}
\end{equation}
where each $\mathbf{T}^{(h)}$ is explicitly aligned with the hour’s variability using $\text{CV}(h)$. This matrix collection serves as the stochastic travel-time "prior" that is later converted into hour-conditioned adjacency weights. This enables the spatial connectivity used by the model to change with the time-of-day rather than remaining fixed across the entire forecasting window.

% -------------------------------------------------------------------

\subsection{Crash Attributes Integration in Adjacency Matrices}
\label{subsec:crash_mapping}
We use the crash data from the Ohio Department of Transportation (ODOT), which are then (along with their attributes) aligned in space and time with the ODOT traffic count dataset. This is done to enable their aggregation into hour-conditioned risk signals used in our adjacency construction. Each crash record $n$ provides a location, timestamp, incident clearance time, followed by contextual attributes like weather/functional-class/work-zone code, vehicle speed, and posted speed limit. Table~\ref{tab:crash_fields} summarizes the notation used throughout the paper for these attributes.

\begin{table}[h]
\centering
\caption{Crash Data and Notations}
\label{tab:crash_fields}
\begin{tabular}{|l|l|l|}
\hline
\textbf{Symbol} & \textbf{Description} & \textbf{Units / Type} \\
\hline
$(\phi_n,\lambda_n)$ & latitude/longitude & degrees \\
\hline
$t_n$ & timestamp & datetime \\
\hline
$C_n$ & incident clearance time & minutes \\
\hline
$\omega_n$ & weather condition code & categorical \\
\hline
$f_n$ & roadway functional class code & categorical \\
\hline
$v_n$ & observed vehicle speed & mph (or km/h) \\
\hline
$v^{\text{lim}}_n$ & posted speed limit & mph (or km/h) \\
\hline
$z_n$ & work-zone indicator & $\{0,1\}$ \\
\hline
\end{tabular}
\end{table}

The first step is to perform the spatial mapping, because crash locations do not coincide exactly with the traffic stations. So, we associate each crash to its nearest station in $\mathcal{V}$:
\begin{equation}
\pi(n) \;=\; \arg\min_{i\in\mathcal{V}} \; d\!\left((\phi_n,\lambda_n),(\phi_i,\lambda_i)\right)
\end{equation}
where $d(\cdot,\cdot)$ denotes a distance metric (e.g., geodesic distance or a local Euclidean approximation). We then further assign each crash to its particular hour-of-day, which aligns the crash aggregation with the hour-conditioned travel-time matrices $\{\mathbf{T}^{(h)}\}_{h=0}^{23}$ defined in Equation \ref{eq:T_bank}.

As our goal is not to treat each attribute as an independent factor in the adjacency matrix, but to build a single combined severity signal that reflects how disruptive a crash can be, which is done by the following formulation:

\noindent\textbf{Incident Clearance Time (ICT) factor}: Let $\bar{C}$ denote the global mean clearance time across the crash dataset. For crash $n$:
\begin{equation}
c_n \;=\; \frac{C_n}{\bar{C}+\varepsilon}.
\end{equation}

\noindent\textbf{Speed violation factor}: We define a non-negative overspeed ratio as:
\begin{equation}
r_n \;=\; \max\!\left(0,\frac{v_n - v_n^{\text{lim}}}{v_n^{\text{lim}}+\varepsilon}\right)
\end{equation}
and normalize it using the global mean $\bar{r}$:
\begin{equation}
\tilde r_n \;=\;
\begin{cases}
\dfrac{r_n}{\bar{r}+\varepsilon}, & \bar{r}>0\\[6pt]
1, & \bar{r}=0
\end{cases}
\end{equation}

\smallskip
\noindent\textbf{Weather and functional-class factors:} By comparing mean clearance times by category, we formulate these factors. Let $\mathbb{E}[C\mid \omega]$ be the mean clearance for weather code $\omega$, and similarly $\mathbb{E}[C\mid f]$ for functional class $f$. Then we define:
\begin{equation}
m_{\omega_n} \;=\; \frac{\mathbb{E}[C\mid \omega_n]}{\bar{C}+\varepsilon},
\qquad
m_{f_n} \;=\; \frac{\mathbb{E}[C\mid f_n]}{\bar{C}+\varepsilon}.
\end{equation}
The conditional means are computed empirically as:
\begin{equation}
\mathbb{E}[C \mid \omega]
\;\approx\;
\frac{1}{N_{\omega}}
\sum_{n:\,\omega} C_n
\end{equation}
\begin{equation}
\mathbb{E}[C \mid f]
\;\approx\;
\frac{1}{N_{f}}
\sum_{n:\,f} C_n
\end{equation}
where $N_{\omega}$ and $N_f$ are the number of crashes in each weather, and functional-class groups.

\smallskip
\noindent\textbf{Work-zone factor:} The work-zone factor $z_n\in\{0,1\}$ is integrated using:
\begin{equation}
m_{z_n} \;=\; \frac{\mathbb{E}[C\mid z_n]}{\bar{C}+\varepsilon},
\end{equation}

\begin{equation}
\mathbb{E}[C \mid z]
\;\approx\;
\frac{1}{N_{z}}
\sum_{n:\,z} C_n
\end{equation}
where $N_z$ denote the number of crashes in work-zone group.

Eventually, we arrive at the proposed definition of crash-level combined severity as:
\begin{equation}
s_n \;=\; c_n \cdot \tilde r_n \cdot m_{\omega_n}\cdot m_{f_n}\cdot m_{z_n}
\label{eq:crash_severity}
\end{equation}

% -------------------------------------------------------
\subsection{Hourly Crash Risk Signals}
\label{subsec:node_edge_risk}

As the crash records are localized data with their associated attributes, and forecasting models usually ingest a pairwise connectivity matrix (a travel-time-based and hour-specific in our case). In order to address this, we need to first aggregate the crash influence at the node level by hour-of-day, and then project it to edge edge-level signal. This modulates the travel-time–based connectivity, as each crash record $n$ is mapped to its nearest station $\pi(n)\in\mathcal{V}$ and has a combined severity score $s_n$. Now, for each hour-of-day $h\in{0,\dots,23}$, we define the node risk as the accumulated severity at station $i$ as:

\begin{equation}
\mathcal{C}_{h,i}=\{\,n \mid \mathrm{hour}(n)=h,\;\pi(n)=i\,\}
\end{equation}
\begin{equation}
\mathrm{R}(h,i)=\sum_{n\in\mathcal{C}_{h,i}} s_n
\label{eq:node_risk_new}
\end{equation}
To make it more clear, $\mathrm{R}(h,i)$ depicts the cumulative impact of a crash near a station $i$ at hour $h$. When no crash maps to $(h,i)$, the risk can be $\mathrm{R}(h, i)=0$, meaning no additional incident affect is applied at that station-hour; this does not imply that the resulting graph connectivity becomes binary. In the subsequent adjacency construction, edge weights remain dense and continuous due to the applied travel-time kernel. Next step is to standardize $\mathrm{R}(h,i)$ as:
\begin{equation}
\widehat{\mathrm{R}}(h,i)=
\frac{\mathrm{R}(h,i)-\mu_{\mathrm{R}}}{\sigma_{\mathrm{R}}+\varepsilon}
\label{eq:risk_standardized_new}
\end{equation}
where $\mu_{\mathrm{R}}$ and $\sigma_{\mathrm{R}}$ are the global mean and standard deviation of $\mathrm{R}(h,i)$, and $\varepsilon>0$ ensures numerical stability. Now, the subsequent graph creation builds on the station pairs $(i,j)$, i.e., dense $N\times N$ interactions used to form an hour-based adjacency. Therefore, we need to project the node-level signal to the pairwise matrix by summing the risks of the endpoints:
\begin{equation}
\widehat{R}_{total}(h,i,j)=\widehat{\mathrm{R}}(h,i)+\widehat{\mathrm{R}}(h,j)
\qquad \forall \quad i,j\in\mathcal{V}
\label{eq:edge_rhat_new}
\end{equation}
The understanding here is that if the endpoint stations observe an increased severity at hour $h$, then the connectivity between the two stations is more likely to be affected (maybe due to localized queues propagating upstream/downstream). Technically, $\widehat{R}_{total}(h,i,j)$ portrays an hour-specific ``incident affect'' situation that can be used to perturb sampled travel times before converting them into the final hour-conditioned adjacency matrix.

% -------------------------------------------------------
\subsection{Formulating Incident-Aware Adjacency Matrix}
\label{subsec:corr_coupling}

After the formulation of $\widehat{R}_{total}(h,i,j)$, we now need to establish the relation between each node pair's $(i,j)$ aggregated crash-context signal and sampled travel times. To accomplish this, we use Pearson correlation coefficients as:
\begin{equation}
\rho_{ij}
= \mathrm{Corr}_h\big(T_{ij}^{(h)},\, \widehat{R}_{\text{total}}(h,i,j)\big),
\label{eq:rho_edge}
\end{equation}
where $\mathrm{Corr}_h(\cdot,\cdot)$ denotes the Pearson correlation computed over the 24 hourly samples $h=0,\dots,23$. And, we bound the correlation coefficient to maintain numerical stability as:
\begin{equation}
\rho_{ij} \leftarrow \mathrm{clip}(\rho_{ij},-\rho_{\max},\rho_{\max}).
\label{eq:rho_clip}
\end{equation}
We then define the incident-aware effective travel time for hour $h$ as:
\begin{equation}
\widehat{T}^{(h)}_{ij}
=
T^{(h)}_{ij}\Big(1 + \rho_{ij}\,\widehat{R}_{total}(h,i,j)\Big),
\qquad (i,j)\in\mathcal{E}
\label{eq:Teff}
\end{equation}
Finally, we convert the incident-aware travel times into hour-conditioned adjacency weights using a Gaussian kernel \cite{STGCN}:
\begin{equation}
A^{(h)}_{ij}
=
\exp\!\left(
-\frac{1}{2\sigma^2}
\left(\frac{\widehat{T}^{(h)}_{ij}}{T^{(h)}_{\max}+\varepsilon}\right)^2
\right)
\label{eq:adj_kernel}
\end{equation}

\begin{equation}
T^{(h)}_{\max}=\max_{(u,v)\in\mathcal{E}}\widehat{T}^{(h)}_{uv}
\end{equation}
\noindent
where \(T^{(h)}_{\max}\) is the maximum effective travel time at hour \(h\), and \(\sigma^2\) controls how sharply connectivity decays with increasing travel time. Finally, we obtain the adaptive adjacency matrix by applying the data-availability mask $\mathbf{A}_{\text{avail}}$ defined in Equation \ref{eq:avail_outer} to the hour-conditioned adjacency:
\begin{equation}
\mathbf{A}^{(h)}_{\text{adaptive}}[i,j]
=
A^{(h)}_{ij}\;\cdot\;\mathbf{A}_{\text{avail}}[i,j]
\label{eq:adj_adaptive}
\end{equation}

This yields a collection of incident-aware, hour-conditioned adaptive adjacency matrices where each $\mathbf{A}^{(h)}_{\text{adaptive}}$ combines incident-aware travel-time perturbations (incident clearance, weather, functional class, speeding, and work-zone context) with the data-availability mask.

% -------------------------------------------------------------------
\subsection{Model Architecture and Adaptive Conformal Prediction (ACP)}
\label{subsec:stt_acp}

We leverage a Spatio-Temporal Transformer model \cite{Xu2020STTN} with an encoder-decoder design (STT-ED) to perform multi-horizon traffic forecasting, exploiting the inherent parallel sequence architecture of a transformer. Let the traffic flow tensor be \(\mathbf{F}\in\mathbb{R}^{T\times N}\), where \(N=|\mathcal{V}|\) sensors and \(T\) is the number of time samples. For each training sample, we form an input window of length \(L\) and predict the next \(H\) steps:
\begin{equation}
\mathbf{X}_t \;=\; \mathbf{F}_{t-L+1:t}\in\mathbb{R}^{L\times N},
\qquad
\mathbf{Y}_t \;=\; \mathbf{F}_{t+1:t+H}\in\mathbb{R}^{H\times N}
\end{equation}
As our spatial information is hour-conditioned, each window is characterized by an hour-of-day index \(h_t\in\{0,\dots,23\}\), extracted from the timestamp of the last step in the window. Now $\mathbf{A}^{(h)}_{\text{adaptive}}$, is injected in the spatial module of the model by selecting the appropriate matrix based on the window hour \(h_t\). Then, a row-normalized version is implemented for graph mixing:
\begin{align}
\widetilde{\mathbf{A}}_t
&=\mathrm{RowNorm}\!\left(\mathbf{A}^{(h_t)}_{\text{adaptive}} + \mathbf{I}\right)
\label{eq:rownormA}\\
\mathrm{RowNorm}(\mathbf{M})[i,j]
&=\frac{\mathbf{M}[i,j]}{\sum_{k}\mathbf{M}[i,k]+\varepsilon}
\label{eq:rownormDef}
\end{align}

\noindent
\textbf{Temporal Tokenization and Temporal Encoder:} Unlike recurrent models, the Transformer processes a sequence of tokens. For each node \(i\), it treats \(L\)-step history as a 1D sequence and converts it into patch tokens using a temporal convolution with patch length \(p\) \cite{Nie2023PatchTST} (e.g., \(p=6\) for 90-min patches with 15-min data). Let \(x_{t,i}\in\mathbb{R}^{L}\) denote the input history for node \(i\). The tokens are formed as:
\begin{equation}
\mathbf{Z}^{(0)}_{t,i}
\;=\;
\mathrm{PatchConv}(x_{t,i})
\;\in\;
\mathbb{R}^{L_p\times d},
\qquad
L_p=\left\lfloor \frac{L}{p}\right\rfloor
\end{equation}
where \(d\) is the Transformer embedding dimension. We add sinusoidal positional encoding \(\mathrm{PE}(\cdot)\) to preserve temporal ordering as:
\begin{equation}
\mathbf{Z}^{(0)}_{t,i} \leftarrow \mathbf{Z}^{(0)}_{t,i} + \mathrm{PE}
\end{equation}
The temporal encoder comprises \(D_t\) stacked Transformer encoder blocks. For a generic encoder layer \(\ell\), multi-head self-attention (MHSA) is computed as:
\begin{align}
\mathbf{Q} &= \mathbf{Z}^{(\ell-1)}\mathbf{W}_Q
\label{eq:Qdef}\\
\mathbf{K} &= \mathbf{Z}^{(\ell-1)}\mathbf{W}_K
\label{eq:Kdef}\\
\mathbf{V} &= \mathbf{Z}^{(\ell-1)}\mathbf{W}_V
\label{eq:Vdef}\\
\mathrm{Attn}(\mathbf{Q},\mathbf{K},\mathbf{V})
&=
\mathrm{softmax}\!\left(\frac{\mathbf{Q}\mathbf{K}^\top}{\sqrt{d_k}}\right)\mathbf{V}
\label{eq:attn}
\end{align}
With residual connections and layer normalization (LN), the temporal encoder update is:
\begin{align}
\widehat{\mathbf{Z}}^{(\ell)} &= \mathrm{LN}\!\Big(\mathbf{Z}^{(\ell-1)}
+ \mathrm{Dropout}(\mathrm{MHSA}(\mathbf{Z}^{(\ell-1)}))\Big)\\
\mathbf{Z}^{(\ell)} &= \mathrm{LN}\!\Big(\widehat{\mathbf{Z}}^{(\ell)}
+ \mathrm{FFN}(\widehat{\mathbf{Z}}^{(\ell)})\Big)
\end{align}
where \(\mathrm{FFN}(\cdot)\) is a two-layer MLP with nonlinearity. The output of the temporal encoder produces per-node temporal memories \(\mathbf{M}^{\text{time}}_{t,i}\in\mathbb{R}^{L_p\times d}\).

\noindent
\textbf{Adjacency Matrix Integration with Spatial Module:} To integrate the hour-conditioned adjacency matrix in the spatial module, temporal tokens are pooled using global average pooling as:
\begin{equation}
\mathbf{u}_{t,i}
\;=\;
\mathrm{Pool}\!\left(\mathbf{M}^{\text{time}}_{t,i}\right)
\in\mathbb{R}^{d}
\end{equation}
We then learn a trainable node embedding \(\mathbf{e}_i\in\mathbb{R}^{d}\) to encode each station identity as:
\begin{equation}
\mathbf{h}_{t,i}^{(0)} \;=\; \mathbf{u}_{t,i} + \mathbf{e}_i
\end{equation}
Now, stacking all the nodes results in \(\mathbf{H}_t^{(0)}\in\mathbb{R}^{N\times d}\). Then, we inject the hour-conditioned adaptive adjacency using:
\begin{equation}
\mathbf{H}_t^{\text{mix}} \;=\; \widetilde{\mathbf{A}}_t\,\mathbf{H}_t^{(0)}
\;\in\;\mathbb{R}^{N\times d}
\label{eq:graph_mix}
\end{equation}
This is the key mechanism by which \(\mathbf{A}^{(h)}\) directly used in the neural representation. Before the spatial attention is applied, the node embeddings are already skewed by the incident/variability-aware connectivity for the corresponding hour-of-day.

\noindent
\textbf{Spatial Transformer Encoder:} After the integration of hour-conditioned adjacency matrix, we refine the spatial interactions using a spatial Transformer encoder consisting of \(D_s\) stacked encoder blocks operating over the nodes. We start the spatial encoder from the graph-mixed representation. That is, the output of the adjacency-based mixing step becomes the spatial encoder’s initial hidden state, \(\mathbf{H}_t^{(0,s)}=\mathbf{H}_t^{\text{mix}}\). This just ensures, the hour-conditioned spatial graphs are included in the spatial tokens before attention is learned. After the spatial encoder, each node embedding is treated as a single spatial token:
\begin{equation}
\mathbf{s}_{t,i} \;=\; \mathbf{H}_t^{(D_s,s)}[i,:] \in \mathbb{R}^{d},
\qquad i=1,\dots,N
\end{equation}

%-----------------------------------------------------------------------------------------
\noindent
\textbf{Long-Horizon Decoding with Cross-Attention:} After the temporal encoder and the spatial encoder, each node \(i\) has temporal memory tokens \(\mathbf{M}^{\text{time}}_{t,i}\), and a spatial token \(\mathbf{s}_{t,i}\). Now, a decoder memory is created by concatenation:
\begin{equation}
\mathbf{M}_{t,i}
=
\Big[\mathbf{M}^{\text{time}}_{t,i}\;\|\;\mathbf{s}_{t,i}\Big]
\in\mathbb{R}^{(L_p+1)\times d}
\label{eq:mem_concat}
\end{equation}
To forecast the next \(H\) steps, we use an \(H\)-length learnable query seed \(\mathbf{Q}_0\in\mathbb{R}^{H\times d}\) (tiled per node) and decode using a standard Transformer decoder.
At each decoder layer, the query first looks at itself (to couple horizons) and then cross-attends to the memory block:
\begin{align}
\mathbf{Q}'_{t,i} &= \mathrm{Attn}(\mathbf{Q}_{t,i},\mathbf{Q}_{t,i},\mathbf{Q}_{t,i})\\
\mathbf{Q}_{t,i} &\leftarrow \mathrm{Attn}(\mathbf{Q}'_{t,i},\mathbf{M}_{t,i},\mathbf{M}_{t,i})
\label{eq:cross_attn_decode}
\end{align}
Finally, the decoded representation is mapped to the \(H\)-step flow forecast for node \(i\) by a linear head:
\begin{equation}
\widehat{\mathbf{Y}}_t[:,i] = \mathbf{Q}_{t,i}\mathbf{W}_o + \mathbf{b}_o,
\qquad
\widehat{\mathbf{Y}}_t\in\mathbb{R}^{H\times N}
\label{eq:decoder_head}
\end{equation}

\noindent
\textbf{Adaptive Conformal Prediction (ACP) for Multi-Horizon Intervals:}
Following our prior work~\cite{Patil2025}, we estimate uncertainty via ACP with epoch-wise adaptive calibration. After each training epoch (and once after early-stopping selects the final checkpoint), we compute calibration residuals on the held-out validation set:
\begin{equation}
r_{t,k,i} \;=\; \left|\,\widehat{Y}_{t,k,i} - Y_{t,k,i}\,\right|,
\end{equation}
and obtain horizon- and node-specific conformal radii as
\begin{equation}
q_{k,i}^{(1-\alpha)}
\;=\;
\mathrm{Quantile}_{1-\alpha}\Big(\{r_{t,k,i}\}_{t\in\mathcal{D}_{\mathrm{cal}}}\Big).
\label{eq:qki}
\end{equation}
The resulting prediction intervals are
\begin{equation}
\widehat{Y}_{t,k,i} - q_{k,i}^{(1-\alpha)}
\;\le\;
Y_{t,k,i}
\;\le\;
\widehat{Y}_{t,k,i} + q_{k,i}^{(1-\alpha)}.
\label{eq:pi}
\end{equation}
Here \(q_{k,i}^{(1-\alpha)}\) is computed per horizon and per node to reflect increasing uncertainty with \(k\) and spatial heterogeneity.

 %----------------------------------------------------------------------------
\begin{table*}[!t]
\centering
\setlength{\tabcolsep}{3pt}
\renewcommand{\arraystretch}{1.15}
\caption{Sample ODOT Crash Dataset}
\label{tab:crash_data}
\begin{tabular}{|p{2.2cm}|p{1.0cm}|p{1.6cm}|p{1.6cm}|p{2.1cm}|p{2.1cm}|p{1.9cm}|p{1.6cm}|p{1.1cm}|}
\hline
\textbf{Crash Date/time} & \textbf{Weather} & \textbf{Latitude} & \textbf{Longitude} &
\textbf{Functional Class} & \textbf{Vehicle Speed (mph)} & \textbf{Speed Limit (mph)} & \textbf{Work Zone} & \textbf{ICT (min)} \\ \hline
11/19/2023 11:43 & 1 & 40.111327 & -83.002978 & 1 & 50 & 65 & N & 67 \\ \hline
05/16/2023 06:50 & 2 & 39.832623 & -82.736814 & 3 & 40 & 60 & N & 115 \\ \hline
06/19/2023 12:13 & 2 & 39.945520 & -81.985450 & 1 & 50 & 50 & Y & 64 \\ \hline
\multicolumn{9}{|c|}{\textbf{...}} \\ \hline
11/22/2023 15:25 & 2 & 39.931881 & -82.907888 & 1 & 75 & 70 & N & 50 \\ \hline
\multicolumn{9}{|p{\dimexpr\textwidth-2\tabcolsep-2\arrayrulewidth\relax}|}{\footnotesize
\textit{Weather code:} 1=Clear, 2=Cloudy, 3=Fog/Smog/Smoke, 4=Rain, 5=Sleet/Hail, 6=Snow, 7=Severe Crosswinds, 8=Blowing Sand/Soil/Dirt/Snow, 9=Freezing Rain/Drizzle, 99=Other/Unknown\par
\textit{Functional class code:} 1-2=Interstates/Freeways, 3=Principal Arterial, 4=Minor Arterial, 5-6=Collector Roads, 7=Local Roads\par
\textit{Work zone:} Y=Yes, N=No} \\ \hline

\end{tabular}
\end{table*}

%--------------------------------------------------------------------------------

\begin{figure}[h!] 
\centering 
\includegraphics[width=\linewidth]{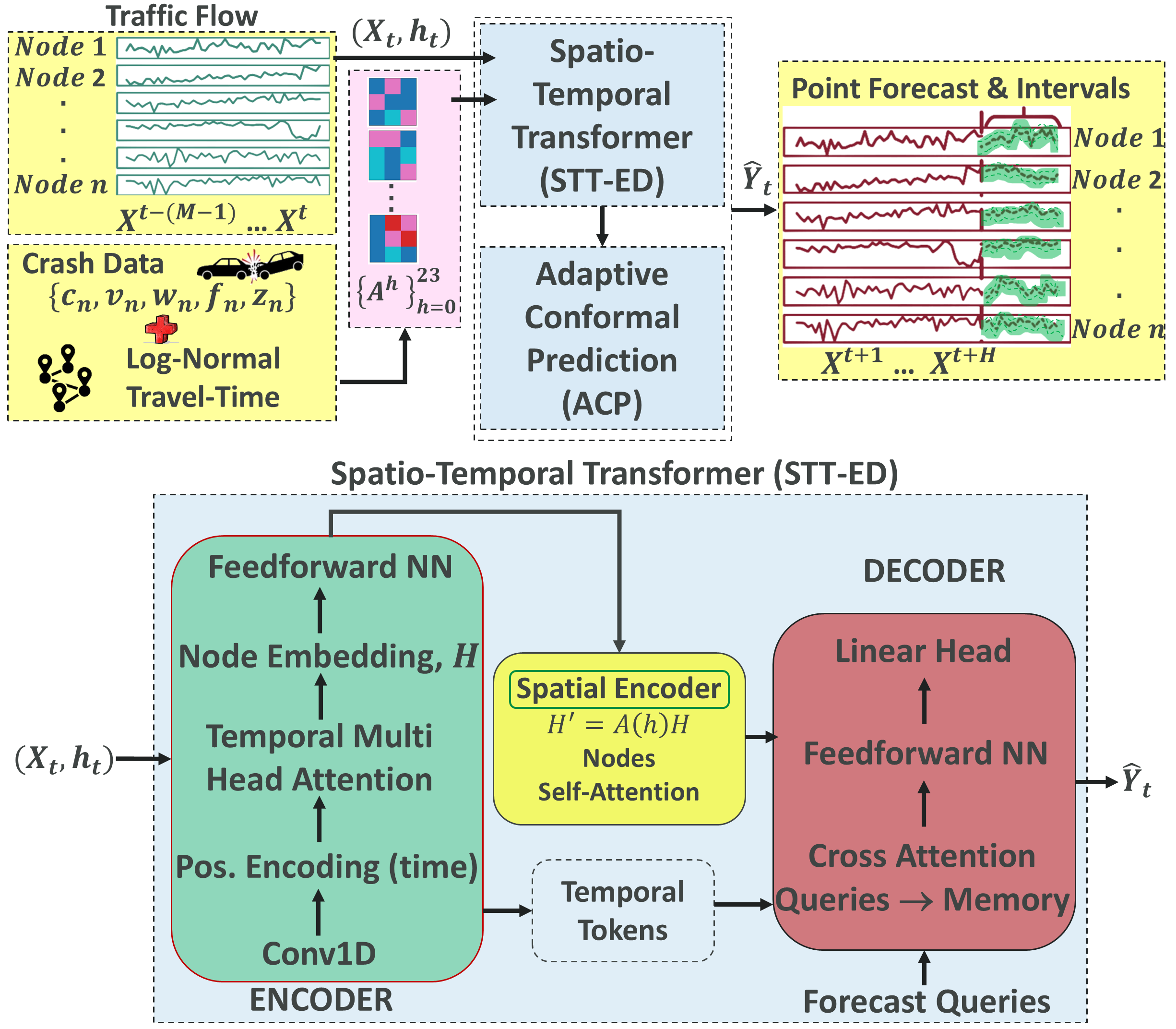} 
\caption{Spatio-Temporal Transformer (STT-ED) architecture with hour-conditioned adaptive adjacency and Adaptive Conformal Prediction (ACP) for multi-horizon traffic forecasting}
\label{fig:model_architecture} 
\end{figure}

Overall architecture of the proposed Spatio-Temporal Transformer with Adaptive Conformal Prediction (STT-ED-ACP) is shown in Figure~\ref{fig:model_architecture}. The STT takes as input a traffic flow window $\mathbf{X}_t$ and the corresponding hour-of-day tag $h_t$, and produces multi-horizon point forecasts $\widehat{\mathbf{Y}}_t$. Crash attributes and baseline travel times are used to construct an hour-conditioned, incident-aware adjacency matrices $\{\mathbf{A}^{(h)}\}_{h=0}^{23}$, from which the appropriate adjacency is selected based on $h_t$ and injected via graph-weighted mixing in the spatial encoder. ACP is applied as a post-forecast calibration layer to generate horizon- and node-specific prediction intervals.

%%%%%%%%%%%%%%%%%%%%%%%%%%%%%%%%%%%%%%%%%%%%%%%%%%%%%%%%%%%%%%%%%%%%%%%%%%%%%%%%%%%%%%%%%%%%%%%%%%%%%

\section{Results and Discussion}
\label{sec:results}

We evaluate the proposed STT-ED framework on 2023 ODOT traffic data and the corresponding ODOT crash records (also reported to the Ohio Department of Public Safety (ODPS)). We set prediction horizons of 1 hr, 2 hr, 3 hr, and 4 hr, with a 24-hour historical look-back period. All models, including the ablation variants, were trained with early stopping (patience = 10) and a maximum of 20 epochs. All experiments are conducted on the full network with \(N=273\) nodes, yielding a \(273\times 273\) travel-time-based adjacency
(\(N^2\) node pairs).
% \begin{figure}[t]
%     \centering
%     \includegraphics[width=\linewidth]{route_matrix1.png}
%     \caption{Study corridor and examples of hour-conditioned adaptive adjacency
%     slices produced by the proposed pipeline (illustrative).}
%     \label{fig:adj_matrices_2023}
% \end{figure}
\noindent
We found 20,046 total crashes in Ohio for 2023. Table~\ref{tab:crash_data} shows a sample of the crash dataset used to build the incident-aware attribute signals. 

\vspace{-2mm}
\subsection{Baseline Methods and Evaluation Metrics}
\label{subsec:baselines_metrics}

We selected the following baseline models for comparative analysis using the same inputs and incident-aware adjacency as the proposed model: % use numbersinstead of bullet- mention that we use same dataset in the models!!!!

\begin{enumerate}
    \item Historical Average (HA) \cite{HA}: Predicts traffic flow using the historical mean computed from the training data.

    \item Autoregressive Integrated Moving Average (ARIMA) \cite{ARIMA}: Classical univariate statistical forecasting model, capturing autoregressive and moving-average dynamics.

    \item Feedforward Neural Network (FNN): Multilayer perceptron capturing non-linear relationships.

    \item GCN-GRU \cite{Fu2016}: Using gated recurrent units (GRU) for temporal modeling to reduce parameter count and improve computational efficiency.

    \item GAT-LSTM \cite{Patil2025}: Using graph attention (GAT) to learn adaptive spatial weights, followed by LSTM-based temporal forecasting.

    \item STGCN~\cite{STGCN}: Spatio-temporal graph convolutional network that jointly models spatial and temporal dependencies using graph convolutions and gated temporal convolutions.

    \item DCRNN~\cite{DCRNN}: Diffusion convolutional recurrent neural network that integrates diffusion-based graph convolutions with a sequence-to-sequence recurrent architecture for multi-step forecasting.

    \item Graph WaveNet~\cite{Wavenet}: Dilated temporal convolutional network augmented with adaptive graph learning and diffusion graph convolutions for long-range spatio-temporal dependencies.

    \item ASTGCN~\cite{ASTGCN}: Attention-based spatio-temporal graph convolutional network that models both spatial and temporal attention mechanisms within a Chebyshev graph convolution framework.
\end{enumerate}

We report standard accuracy metrics for multi-horizon prediction, aggregated over all nodes and forecast steps. Mean Absolute Error (MAE) is:
\begin{equation}
\mathrm{MAE} = \frac{1}{|\Omega|}\sum_{(t,k,i)\in\Omega}
\left|Y_{t,k,i} - \widehat{Y}_{t,k,i}\right|,
\end{equation}
and Root Mean Squared Error (RMSE) is:
\begin{equation}
\mathrm{RMSE} = \sqrt{\frac{1}{|\Omega|}\sum_{(t,k,i)\in\Omega}
\left(Y_{t,k,i} - \widehat{Y}_{t,k,i}\right)^2},
\end{equation}
where \(\Omega\) denotes the evaluation set over time indices \(t\), horizons
\(k\in\{1,\dots,H\}\), and nodes \(i\in\{1,\dots,N\}\).

To evaluate calibrated uncertainty, we report Prediction Interval Coverage Probability (PICP) and Mean Prediction Interval Width (MPIW) using ACP intervals:
\begin{equation}
\mathrm{PICP} = \frac{1}{|\Omega|}\sum_{(t,k,i)\in\Omega}
\mathbf{1}\!\left(\widehat{Y}^{L}_{t,k,i}\le Y_{t,k,i}\le \widehat{Y}^{U}_{t,k,i}\right),
\end{equation}
\begin{equation}
\mathrm{MPIW} = \frac{1}{|\Omega|}\sum_{(t,k,i)\in\Omega}
\left(\widehat{Y}^{U}_{t,k,i}-\widehat{Y}^{L}_{t,k,i}\right),
\end{equation}
where \([\widehat{Y}^{L}_{t,k,i},\widehat{Y}^{U}_{t,k,i}]\) is the conformal prediction interval at horizon \(k\) and node \(i\). As in our prior work, we omit MAPE because near-zero flows can inflate or render percentage errors undefined, while MAE/RMSE remain stable.

\subsection{Adjacency Analysis and Performance Comparison} % talk about why not interstae travels and why doing loops - lack of data, so this is teh way
\label{subsec:perf_adj_analysis}

\begin{figure}[h!] 
\centering 
\includegraphics[width=\linewidth]{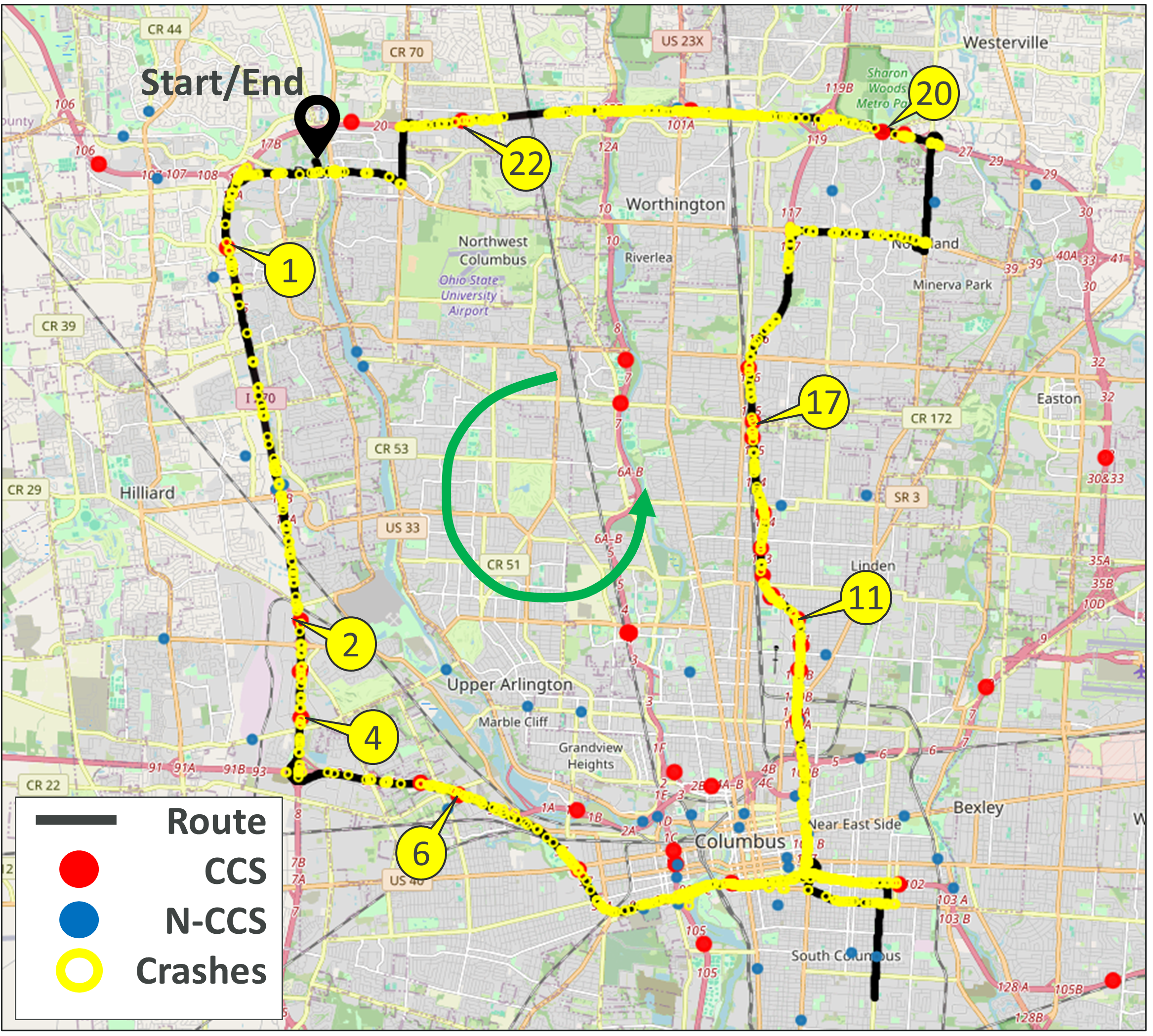} 
\caption{Sampled Traffic Route}
\label{fig:sample_route} 
\end{figure}

To ensure a controlled and interpretable comparison across all models, we evaluate performance on a fixed subset of $N=50$ representative stations, corresponding to a connected route segment within the study network. Figure~\ref{fig:sample_route} illustrates the selected route, traffic count stations, and crash sites. We select this particular Columbus route segment as the length of the route is approximately 47.5 miles (typically varying from 1 hour to 1 hour 30 minutes), which enables repeatable long-run experiments without requiring a much larger corridor-scale network. While a natural alternative would be to validate long-horizon forecasting on a single 4+ hour interstate corridor traversal, doing so would require continuous multi-state data coverage and consistent incident/ground-truth alignment across jurisdictions. Since our study is based on Ohio-only datasets (ODOT traffic counts and ODOT crash records), we adopt a loop-based experimental design on a Columbus corridor to emulate extended trips while remaining fully supported by the available data sources. Also, since our objective is long-horizon forecasting, we require extended scenarios to evaluate how prediction quality evolves as conditions transition across the day. Therefore, we are imitating a travel that executes ``four'' repeated loop runs over the same segment, which effectively spans off-peak to peak (or peak to off-peak) regimes within a single experiment. This design provides a consistent and efficient way to generate long-duration test scenarios while preserving identical spatial structure, and it enables a controlled validation task while undertaking SUMO Monte Carlo simulations.

\begin{figure}[h!] 
\centering 
\includegraphics[width=\linewidth]{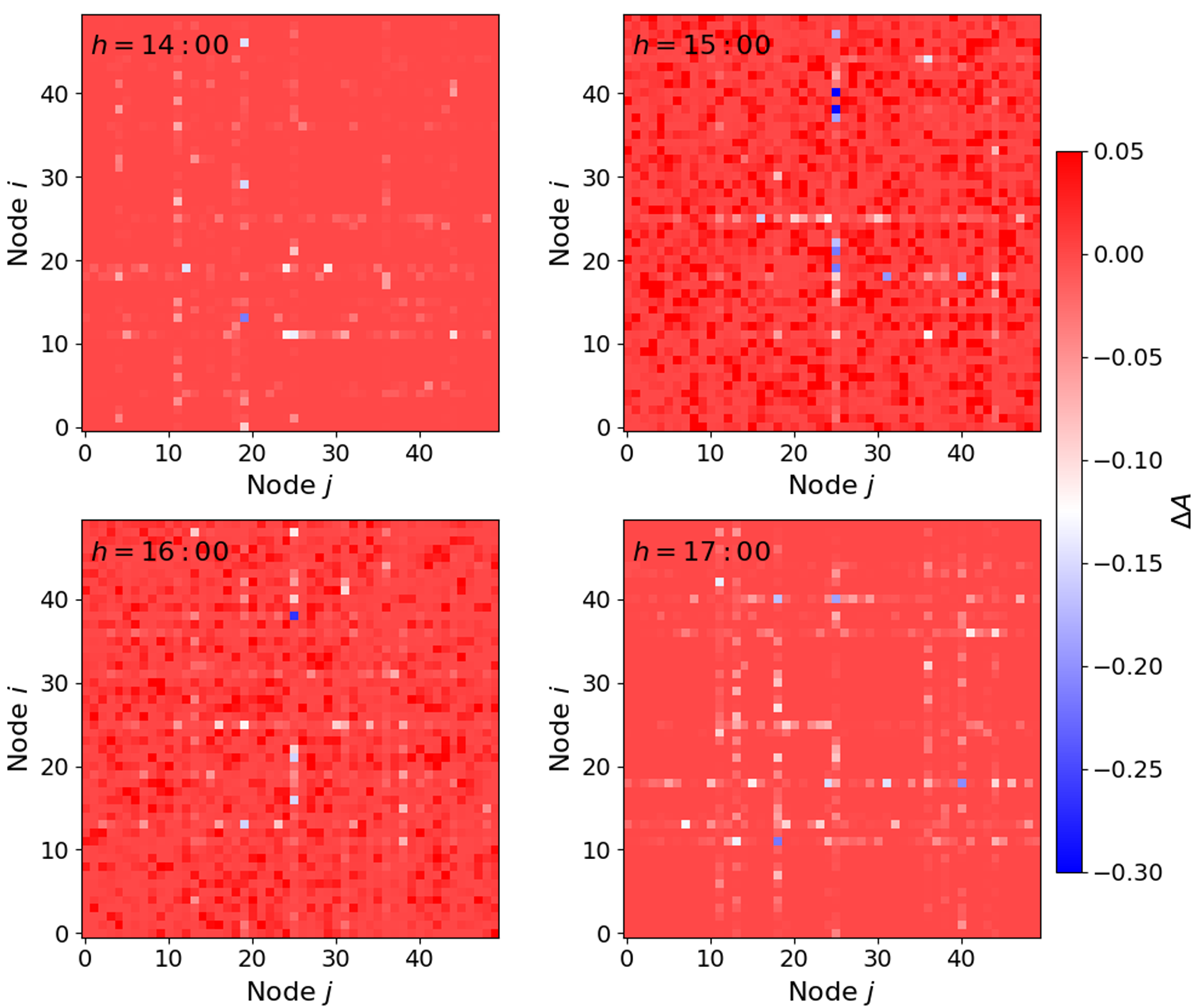} 
\caption{Crash-induced change in adjacency weights across selected hours}
\label{fig:deltaA_hours} 
\end{figure}

Beyond predictive accuracy, we analyze how incorporating crash context alters the underlying spatial connectivity used by the forecasting models. Figure~\ref{fig:deltaA_hours} visualizes the change in adjacency weights:
\[
\Delta \mathbf{A}^{(h)} = \mathbf{A}^{(h)}_{\text{crash}} - \mathbf{A}^{(h)}_{\text{base}}
\]
for four afternoon hours ($h=14:00$ - $17:00$), where $\mathbf{A}^{(h)}_{\text{base}}$ is constructed using CV$(h)$ travel-time sampling, and $\mathbf{A}^{(h)}_{\text{crash}}$ further incorporates crash severity. The predominantly negative $\Delta \mathbf{A}^{(h)}$ values indicate that crash-induced increases in effective travel time reduce graph connectivity through the travel-time kernel, thereby weakening spatial coupling between affected nodes. Importantly, the magnitude and spatial pattern of these changes vary by hour, highlighting the time-dependent impact of incidents on the network. 

Table~\ref{tab:mae_rmse} presents the MAE and RMSE for each method across multi-hour prediction windows.

\begin{table}[!h]
\caption{Prediction Performance Comparison\label{tab:mae_rmse}}
\centering
\setlength{\tabcolsep}{8pt}
\renewcommand{\arraystretch}{1.1}
\begin{tabular}{|m{1.8cm}|m{2.6cm}|m{2.6cm}|}
\hline
\multirow{2}{*}{\textbf{Method}} &
\multicolumn{1}{c|}{\textbf{MAE}} &
\multicolumn{1}{c|}{\textbf{RMSE}} \\
\cline{2-3}
& \textbf{1h / 2h / 3h / 4h} & \textbf{1h / 2h / 3h / 4h} \\
\hline
HA & 0.921/0.921/0.921/0.921 & 0.991/0.991/0.991/0.991 \\ \hline
ARIMA & 0.136/0.327/0.520/0.674 & 0.191/0.413/0.608/0.752 \\ \hline
FNN & 0.357/0.380/0.389/0.392 & 0.511/0.558/0.569/0.571 \\ \hline
GCN-GRU & 0.310/0.326/0.347/0.401 & 0.394/0.426/0.490/0.511 \\ \hline
GAT-LSTM & 0.251/0.265/0.362/0.426 & 0.363/0.370/0.477/0.514 \\ \hline
STGCN & 0.321/0.323/0.335/0.359 & 0.418/0.419/0.462/0.472 \\ \hline
DCRNN & 0.098/0.227/0.360/0.486 & 0.148/0.321/0.474/0.614 \\ \hline
Graph WaveNet & 0.122/0.233/0.350/0.485 & 0.157/0.282/0.422/0.601 \\ \hline
ASTGCN & 0.128/0.228/0.329/0.332 & 0.162/0.311/0.432/0.461 \\ \hline
\textbf{STT-ED} & \textbf{0.087/0.214/0.291/0.317} & \textbf{0.138/0.277/0.374/0.413} \\ \hline
\end{tabular}
\end{table}

It can be observed that the strongest competitors to STT-ED are the graph-based baselines (notably DCRNN, Graph WaveNet, ASTGCN, and GAT-LSTM), which consistently outperform classical approaches by leveraging spatial correlations across stations. However, most of these methods exhibit larger error growth as the horizon increases, indicating sensitivity to long-horizon uncertainty propagation and limited temporal modeling capacity under distribution shifts. In contrast, STT-ED achieves the best MAE/RMSE across all horizons and shows the most stable scaling with horizon length, suggesting that the encoder-decoder Transformer structure (with cross-attention decoding) is more effective for long-horizon forecasting.

\subsection{Prediction Results}

\begin{figure}[h] \centering \includegraphics[width=\linewidth]{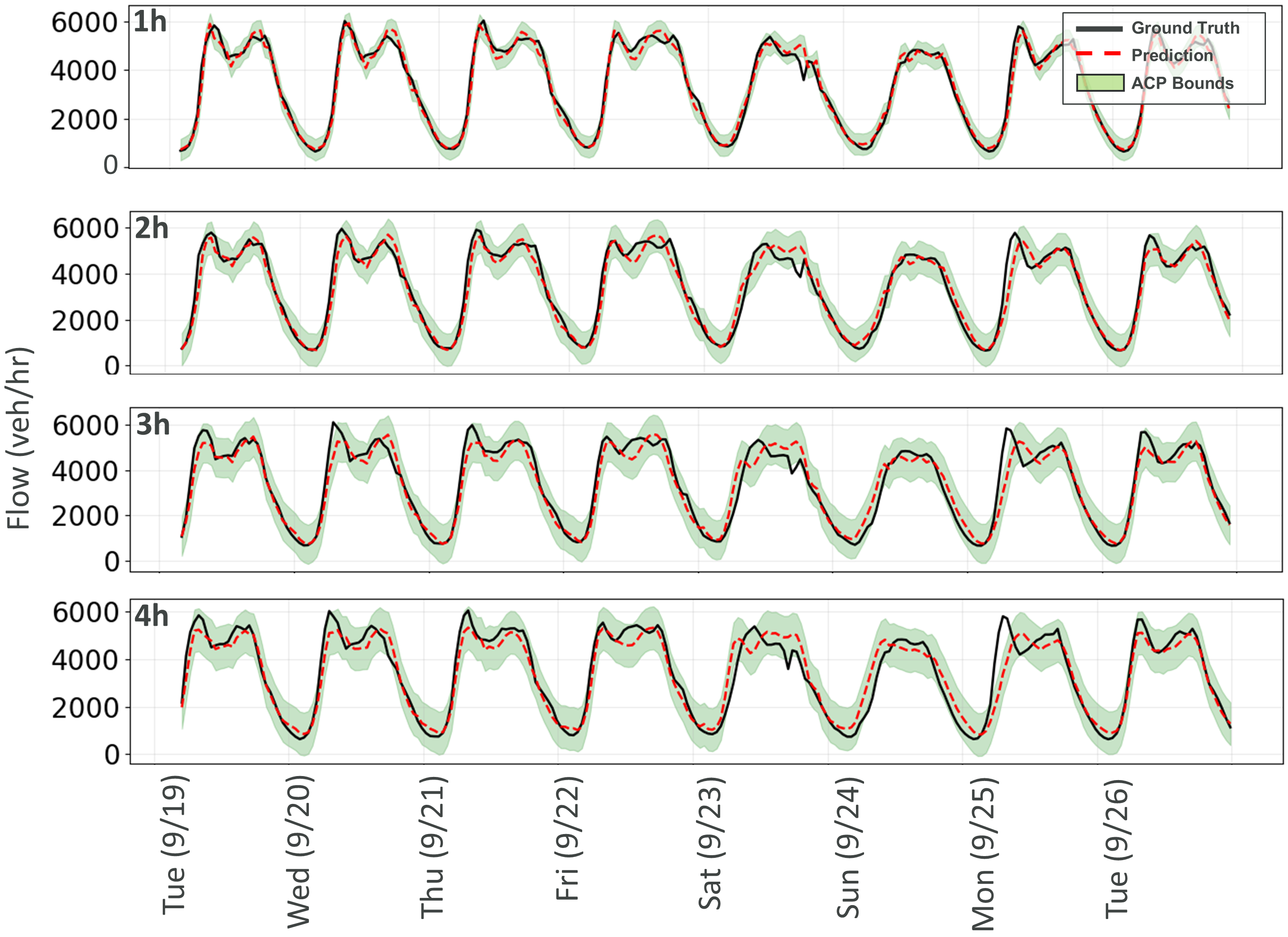} \caption{Traffic Prediction with Uncertainty Bounds (Node 1)} \label{fig:pred} \end{figure}

Figure~\ref{fig:pred} illustrates multi-horizon traffic flow predictions and corresponding ACP uncertainty intervals for Node 1 (see  Figure~\ref{fig:sample_route}). As the forecast horizon increases from 1 to 4 hours, the prediction uncertainty widens, reflecting the accumulation of temporal uncertainty, while the point forecasts continue to capture the dominant patterns and peak–off-peak transitions. Notably, the ACP intervals expand around periods of higher variability, indicating effective uncertainty calibration without overly conservative bounds.

\subsection{Uncertainty Quantification}

To evaluate the effectiveness of the proposed Adaptive Conformal Prediction (ACP) framework, we benchmark it against the following baselines:

\begin{enumerate}
    \item Gaussian Process Regression (GPR) \cite{GPR}: Probabilistic regression framework that yields predictive distributions by utilizing kernel-based Bayesian formulation.   
    \item Mean-Variance Estimation (MVE) \cite{MVE}: Jointly predicts the conditional mean and variance, enabling uncertainty-aware point forecasts.  
    \item Deep Ensembles (DE) \cite{Lakshminarayanan2017DeepEnsembles}: Quantifying uncertainty by aggregating predictions from multiple independently trained neural networks.  
    \item Quantile Regression (QR) \cite{QR}: Estimates conditional quantiles, allowing prediction intervals to be constructed without distributional assumptions.
    \item Bootstrap Aggregation (BA) \cite{BA}: Estimates predictive variability by training multiple models on resampled training data.
    \item Conformal Prediction (CP) \cite{Stankeviciute2021ConformalTS}: A distribution-free uncertainty quantification method constructing prediction intervals using residuals from calibration set.
    \item Conformalized Quantile Regression (CQR) \cite{Romano2019CQR}: Combines quantile regression with conformal calibration to produce adaptive prediction intervals.
\end{enumerate}

These methods serve as strong baselines for assessing calibration quality and interval efficiency of the proposed ACP approach under multi-horizon traffic forecasting settings. Table~\ref{tab:picp_mpiw} reports PICP and MPIW metrics.

\begin{table}[!h]
\caption{Uncertainty Quantification Bounds Comparison\label{tab:picp_mpiw}}
\centering
\setlength{\tabcolsep}{8pt}
\renewcommand{\arraystretch}{1.1}
\begin{tabular}{|m{1cm}|m{2.9cm}|m{2.9cm}|}
\hline
\multirow{2}{*}{\textbf{Method}} &
\multicolumn{1}{c|}{\textbf{PICP \%}} &
\multicolumn{1}{c|}{\textbf{MPIW}} \\
\cline{2-3}
& \textbf{1h / 2h / 3h / 4h} & \textbf{1h / 2h / 3h / 4h} \\
\hline
GPR & 66.64/89.65/88.20/90.11 & 0.372/3.298/3.303/3.296 \\ \hline
MVE & 89.34/90.51/91.02/91.44 & 0.952/1.221/2.565/2.747 \\ \hline
DE & 71.03/73.10/75.65/77.81 & 0.701/0.882/1.034/1.175 \\ \hline
QR & 67.76/66.32/65.57/66.06 & 0.504/0.655/0.768/0.832 \\ \hline
BA & 43.74/37.22/36.74/35.50 & 0.226/0.246/0.251/0.258 \\ \hline
CP & 90.10/91.50/89.38/89.44 & 0.571/1.910/2.231/2.895 \\ \hline
CQR & 91.95/91.47/90.87/90.76 & 0.668/1.971/1.896/1.458 \\ \hline
ACP & \textbf{94.09/93.51/91.13/92.94} & \textbf{0.582/0.876/1.036/1.148} \\ \hline
\end{tabular}
\end{table}

ACP provides the best coverage among the strongest competitors (CP and CQR). Compared to CP, ACP maintains similar (and often slightly higher) PICP while producing substantially tighter intervals, with MPIW reductions of roughly $\sim$53-60\% for the 2h-4h horizons. Relative to CQR, ACP achieves comparable coverage but notably narrower bounds at 2h-3h (about $\sim$45-55\% smaller MPIW), while remaining competitive at 4h. Overall, ACP delivers higher coverage while maintaining the tightest (or near-tightest) uncertainty bands among the conformal baselines.

\subsection{Ablation Study}
To isolate the contribution of each architectural block in the proposed STT framework, we perform an ablation study by removing components from the full encoder-decoder model. In particular, we compare five configurations:
\begin{enumerate}
    \item STT-ED (Full): the complete encoder-decoder Transformer used in our main experiments.
    \item STT-E (Encoder-only): removes the decoder cross-attention and directly maps learned spatio-temporal embeddings to multi-horizon forecasts.
    \item STT-D (Decoder-only): removes the encoder stack and relies on learned horizon queries with a lightweight memory representation to generate multi-step predictions.
    \item STT-ED-NoPatch: removes the temporal patch tokenization (Conv1D tokenization) and instead uses step-wise tokenization, testing the effect of patch-based temporal embedding.
    \item STT-ED-NoCrossAttn: removes the decoder cross-attention to the encoder memory, testing whether explicit encoder--decoder information flow is necessary for long-horizon decoding. 
\end{enumerate}
Table~\ref{tab:ablation_study} reports MAE/RMSE across the multi-hour horizons.

\begin{table}[h]
\caption{Ablation Study under Different STT Configurations\label{tab:ablation_study}}
\centering
\setlength{\tabcolsep}{6pt}
\renewcommand{\arraystretch}{1.1}
\begin{tabular}{|m{1.7cm}|m{2.9cm}|m{2.9cm}|}
\hline
\multirow{2}{*}{\textbf{Configuration}} &
\multicolumn{1}{c|}{\textbf{MAE}} &
\multicolumn{1}{c|}{\textbf{RMSE}} \\
\cline{2-3}
& \textbf{1h / 2h / 3h / 4h}
& \textbf{1h / 2h / 3h / 4h} \\
\hline
\textbf{STT-ED} & \textbf{0.087/0.214/0.291/0.317} & \textbf{0.138/0.277/0.374/0.413} \\ \hline
STT-ED-NoPatch           & 0.101/0.238/0.304/0.323 & 0.142/0.297/0.382/0.461 \\ \hline
STT-ED-NoCrossAttn           & 0.736/0.720/0.719/0.722 & 0.773/0.789/0.786/0.791 \\ \hline
STT-E          & 0.096/0.221/0.303/0.327 & 0.151/0.289/0.386/0.425 \\ \hline
STT-D            & 0.137/0.249/0.297/0.322 & 0.178/0.351/0.383/0.421 \\ \hline
\end{tabular}
\end{table}

\noindent
The ablation findings confirm that the full STT-ED achieves the best accuracy and degrades gracefully with horizon, indicating that both the encoder memory and decoder cross-attention are key for stable long-horizon decoding. Removing patch tokenization (STT-ED-NoPatch) produces a consistent error increase across horizons from 1h to 4h, suggesting that temporal patching mainly improves representation efficiency and robustness at longer horizons. The encoder-only variant (STT-E) remains close to the full model (MAE within +0.009 to +0.012; RMSE within +0.013 to +0.019), while the decoder-only variant (STT-D) is notably worse at 1h–2h (MAE +0.050/+0.035; RMSE +0.040/+0.074) but becomes competitive by 4h, implying that the encoder’s temporal token memory is particularly beneficial for near-term dynamics. In contrast, removing decoder cross-attention (STT-ED-NoCrossAttn) causes a severe collapse (MAE $\approx$0.72–0.74; RMSE $\approx$0.77–0.79), demonstrating that explicit encoder-to-decoder information flow is essential for producing meaningful multi-horizon forecasts in this architecture.

\subsection{Model Validation with Multi-Hour Simulation}
\label{subsec:simulation_validation}

To validate the proposed forecasting framework under realistic, time-varying conditions, we extend the simulation-based evaluation strategy used in our prior work~\cite{PatilSUMO} to a multi-hour setting. Unlike single-hour validation, this experiment explicitly assesses the model’s ability to propagate traffic dynamics as network demand evolves across consecutive hours.

Using the same sampled route described in Section~\ref{subsec:perf_adj_analysis}, we conduct microscopic traffic simulations in SUMO driven by the model’s predicted traffic flows for the selected time periods. For each experiment, the STT produces hourly forecasts for all stations along the route, and these predicted flows are injected into SUMO to construct a four-hour traffic scenario. As a result, network demand is updated at each simulated hour, enabling evaluation under realistic transitions between off-peak and peak conditions. The route characteristics are outlined in Table~\ref{tab:route_characteristics}.

\begin{table}[h]
\caption{Route Characteristics\label{tab:route_characteristics}}
\centering
\setlength{\tabcolsep}{10pt}
\begin{tabular}{|m{3.7cm}|m{3.7cm}|}
\hline
\textbf{Parameter} & \textbf{Value} \\
\hline
Route Length (mi) & 47.5 \\
\hline
Number of Intersections & 84 \\
\hline
Number of Traffic Lights & 39 \\
\hline
Number of Lanes (per direction) & 1-5 \\
\hline
Functional Class & Interstates/freeways, arterials, collectors and locals \\
\hline
Mean Speed (mph) & 74.5 (TOD range: 63.2-80.0) \\
\hline
INRIX congestion indicator (\%) & 95.9 (Historic average: 96.3) \\
\hline
\end{tabular}
\vspace{-10pt}
\end{table}

\begin{figure}[!h]
    \centering
    \includegraphics[width=\linewidth]{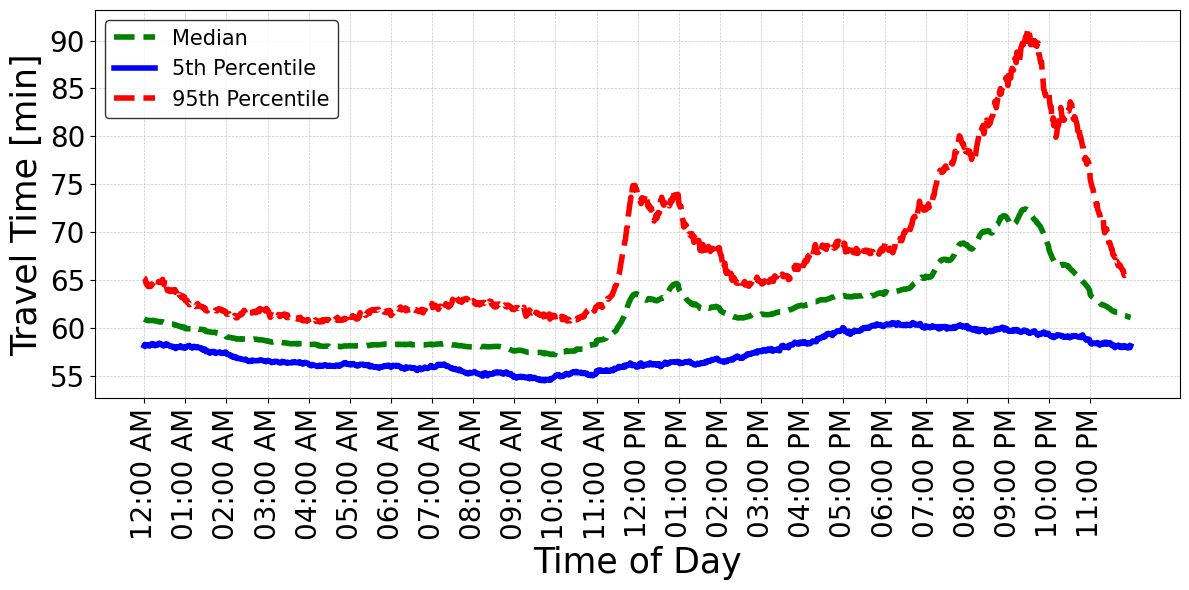}
    \caption{INRIX historical travel time data}
    \label{fig:inrixData}
\end{figure}

We focus on two representative peak regimes as shown in Figure~\ref{fig:inrixData} as observed in historical INRIX travel-time data. There is a daytime peak from 10:00 AM to 2:00 PM and an evening peak from 6:00 PM to 10:00 PM.  The INRIX data represents single-trip travel times for the selected 47.5-mile route, aggregated by departure hour (i.e., one traversal of the corridor). For each scenario, we simulate a 200-run Monte Carlo simulation to cover the variability pattern.

\begin{figure}[!h]
    \centering

    \subfloat[12:00 (daytime)\label{fig:inrixHeatmapNoon}]{
        \includegraphics[width=\linewidth]{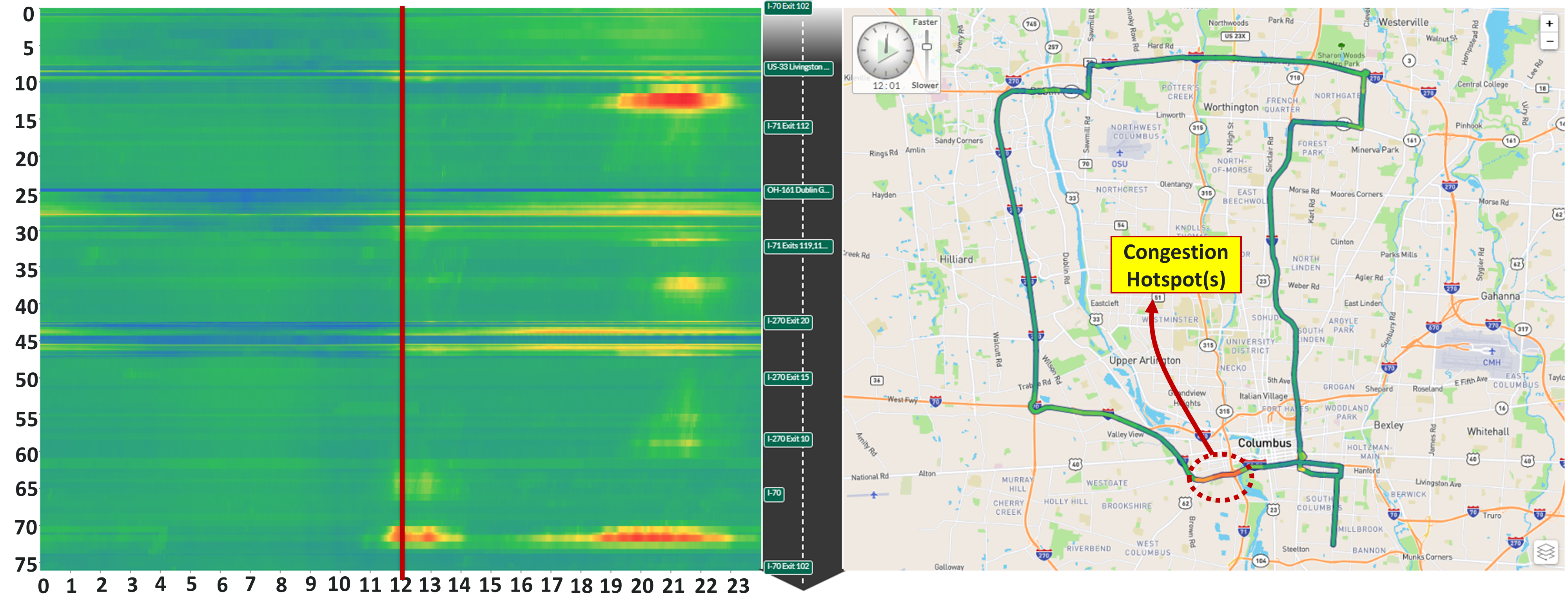}
    }

    \vspace{0.3em}

    \subfloat[21:00 (evening)\label{fig:inrixHeatmapNight}]{
        \includegraphics[width=\linewidth]{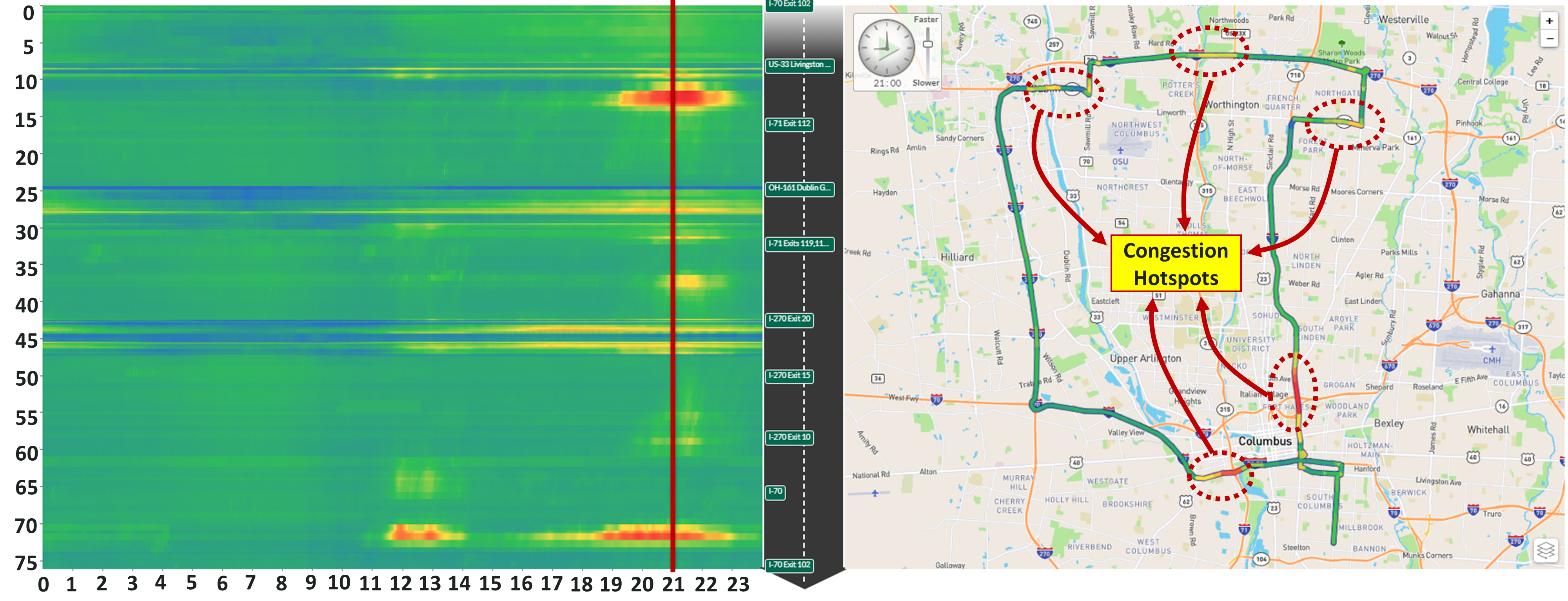}
    }

    \caption{INRIX Congestion Heatmaps}
    \label{fig:inrixHeatmaps}
\end{figure}

Figure~\ref{fig:inrixHeatmaps} shows INRIX corridor heatmaps for two representative departure times (12:00 and 21:00), illustrating the spatial heterogeneity of congestion along the route. We use these plots as qualitative context to identify recurrent hotspots and support the selection of peak regimes, rather than as direct simulation ground truth.

\begin{figure}[!h] 
\centering 
\includegraphics[width=\linewidth]{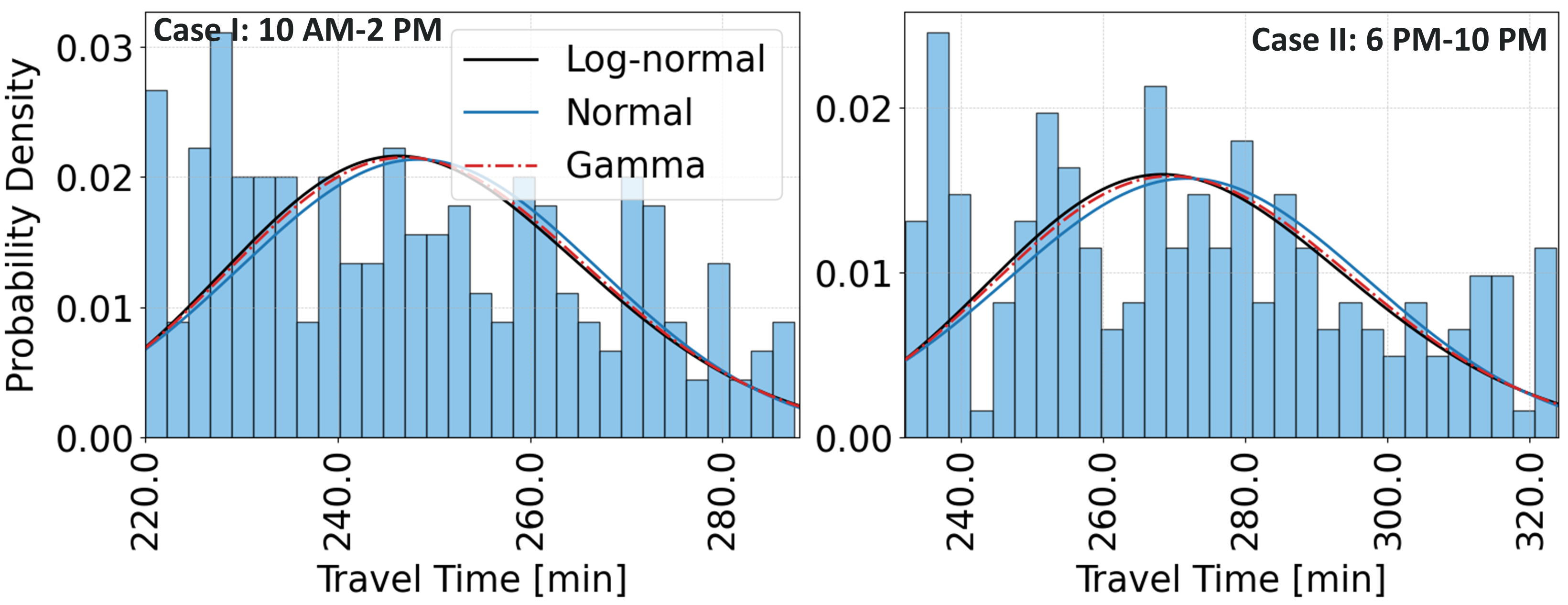} 
\caption{SUMO Travel time distribution (Simulated)} 
\label{fig:validation} 
\end{figure}

At each Monte Carlo run, a Vehicle Under Test (VUT) navigates the route ``four" times; each run consists of multiple sequential traversals, during which traffic flow values are updated hourly based on the model’s forecasts. Figure~\ref{fig:validation} compares the resulting simulated travel-time distributions for the two peak regimes, which explains the shift in the distribution support (approximately 220-330 minutes) relative to the INRIX single-trip statistics (approximately 55-85 minutes). The observed widening of the evening-peak distribution is consistent with the compounding effects of congestion and stop-and-go dynamics across successive peak hours.

Kolmogorov–Smirnov (KS) tests further confirm that the simulated travel-time samples follow a log-normal distribution in both cases, validating the distributional assumption used in the hour-conditioned adjacency formulation.

\begin{itemize}
  \item Case I (Daytime Peak): log-normal (KS = 0.078/p = 0.156), normal (KS = 0.088/p = 0.081), gamma (KS = 0.0809/p = 0.137)
  \item Case II (Evening Peak): log-normal (KS = 0.066/p = 0.333), normal (KS = 0.083/p = 0.116), gamma (KS = 0.071/p = 0.245)
\end{itemize}

Lower KS and \(p{>}0.05\) for log-normal validate both the assumed distribution and simulation fidelity.

\section{Conclusion}
\label{sec:conclusion}

In this work, we proposed an STT-ED forecasting framework with Adaptive Conformal Prediction (STT-ED-ACP) for long-horizon traffic prediction under time-varying and incident-driven conditions. The key idea is to replace static spatial connections with an hour-conditioned, incident-aware adjacency matrix by parameterizing travel-time variability using a piecewise CV$(h)$ profile (log-normal sampling) and perturbing edge weights using crash-derived severity signals. The encoder-decoder Transformer leverages parallel temporal attention and cross-attention decoding, while the selected $\mathbf{A}^{(h_t)}$ is injected into the spatial encoder to reflect prevailing network conditions within each forecasting window. Across all multi-hour horizons, STT-ED achieves the best MAE/RMSE with slower error growth, and ACP provides near-nominal coverage with tight, horizon-dependent prediction intervals, especially during regime transitions. Finally, multi-hour SUMO loop validation shows realistic travel-time distributions consistent with INRIX observations, supporting both the log-normal travel-time assumption and the model's ability to propagate network dynamics over extended horizons.

Despite these encouraging results, several limitations remain. First, the hour-conditioned CV profile is estimated from historical data and treated as fixed during inference; while this captures systematic diurnal patterns, it does not adapt online to unexpected demand surges or large-scale events. Second, crash information is incorporated through aggregated severity signals rather than through explicit spatio-temporal incident evolution, which may oversimplify complex queue formation and dissipation processes. Third, the current study focuses on a single regional network and a fixed set of horizons; broader generalization across cities, seasons, and sensing resolutions remains to be explored. Finally, ACP calibration is performed epoch-wise rather than fully online, which may limit responsiveness under abrupt distribution shifts.

Future work will focus on extending to scaling the approach to larger networks and longer horizons, as well as integrating the framework into real-time decision-support pipelines for traffic management and traveler information systems, also remain important avenues for continued research.

% \newpage

 
\vspace{-30pt}

\end{document}